\documentclass[12pt,reqno]{amsart}
\usepackage{latexsym, amsmath, amscd, amssymb, amsthm}
\usepackage{bm}
\usepackage{mathrsfs}   
\usepackage{dsfont}
\usepackage{xspace}
\usepackage{natbib}
\usepackage{setspace}
\usepackage{enumerate}
\usepackage{verbatim}
\usepackage[english]{babel}
\usepackage{epsfig}
\usepackage{algorithmic,algorithm}

\newtheorem{lemma}{Lemma}[section]
\newtheorem{theorem}[lemma]{Theorem}
\newtheorem{corollary}[lemma]{Corollary}

\theoremstyle{definition}

\newtheorem*{remark}{Remark}

\theoremstyle{remark}

\newcommand{\ie}{{\em i.e.},\xspace }

\newcommand{\N}{\mathbb{N}}
\newcommand{\R}{\mathbb{R}}
\renewcommand{\P}{\mathsf{P}}
\newcommand{\E}{\mathsf{E}}
\newcommand{\Xc}{\mathcal{X}}
\newcommand{\Vc}{\mathcal{V}}

\newcommand{\Ec}{\mathcal{E}}
\newcommand{\Lc}{\mathcal{L}}
\newcommand{\Ac}{\mathcal{A}}
\newcommand{\Dc}{\mathcal{D}}
\newcommand{\Dt}{\widetilde{\Dc}}

\newcommand{\Jt}{\widetilde{J}}
\newcommand{\Lh}{\widehat{L}}

\newcommand{\err}{\mathsf{Err}(\psi,\psi')}
\newcommand{\degr}{\textup{deg}}

\newcommand{\vol}{\textup{vol}}
\newcommand{\var}{\mathsf{Var}}
\newcommand{\1}{\mathds{1}}
\newcommand{\bull}{\cdot}

\begin{document}
\title[Consistency of Spectral Hypergraph Partitioning]
{Consistency of Spectral Hypergraph Partitioning under Planted Partition Model}
\author[D. Ghoshdastidar and A. Dukkipati]{Debarghya Ghoshdastidar, Ambedkar Dukkipati}
\address{\newline Department of Computer Science \& Automation, \newline 
Indian Institute of Science, \newline Bangalore - 560012, India.}
\email{\{debarghya.g,ad\}@csa.iisc.ernet.in}

\begin{abstract}
Hypergraph partitioning lies at the heart of a number of problems 
in machine learning and network sciences.
Many algorithms for hypergraph partitioning have been proposed
that extend standard approaches for graph partitioning to the case of
hypergraphs. However, theoretical aspects of such methods have seldom
received attention in the  literature as compared to the extensive
studies on the guarantees of graph partitioning.  For instance,
consistency results of spectral graph partitioning under the
 stochastic block model are well known.  
In this paper, we present a planted partition model for sparse random  
non-uniform hypergraphs that generalizes the stochastic block model. We derive an error
bound for a spectral hypergraph partitioning algorithm under this model 
using matrix concentration inequalities. 
To the best of our knowledge, this is the first consistency
result related to partitioning non-uniform hypergraphs. 
\end{abstract}

\clearpage\maketitle
\thispagestyle{empty}

\section{Introduction}
A wide variety of complex real-world systems can be understood by analyzing the 
interactions among various entities or components of the system. 
This has made network analysis a subject of both theoretical and practical interest.
A plethora of challenging problems related to social, biological, communication 
networks have intrigued researchers over the past decades, and has led to the 
development of some sophisticated techniques for network analysis. 
This is clearly witnessed in the problems related to network 
or graph partitioning, where the task is to find strongly connected groups of nodes 
with sparse connections across groups.
The problem appears in several engineering applications such as  
circuit or program segmentation~\citep{Kernighan_1970_jour_Bell},
community detection in social or biological networks~\citep{Wasserman_1994_book_Cambridge,Guimera_2005_jour_Nature},
data analysis and clustering~\citep{Ng_2002_conf_NIPS}
among others.  

\subsection*{Graph partitioning and the stochastic block model}
 The problem of finding a balanced partition of a graph is known to be computationally hard.
However, a number of approximate methods have been studied in the literature. These include 
spectral algorithms~\citep{Fiedler_1973_jour_CzechMathJ,Ng_2002_conf_NIPS,Krzakala_2013_jour_PNAS},
modularity and likelihood based methods~\citep{Girvan_2002_jour_PNAS,Bickel_2009_jour_PNAS,Choi_2012_jour_Biometrika},
convex optimization~\citep{Amini_2014_arxiv,Chen_2014_jour_TIT},
belief propagation~\citep{Decelle_2011_jour_PhyRevE} among others.
The empirical success of such methods is not a mere coincidence,
and theoretical guarantees for most of these methods have been extensively studied.
In this respect, it is quite common to study partitioning algorithms under statistical models for
random networks, such as the stochastic block model or planted partition model~\citep{Holland_1983_jour_SocialNetworks} or its variants.
In this model, one considers a random graph on $n$ nodes with a well defined $k$-way
partition $\psi:\{1,\ldots,n\}\to\{1,\ldots,k\}$.
The edges are randomly added with probabilities depending on the class labels
of the participating nodes. 
Thus, the following interesting question arises: 
\vskip0.5ex
\noindent {\it Question:} 
Let $\psi'$ be the partition obtained from an algorithm, then what is the number of mismatches
between $\psi$ and $\psi'$? 
\vskip0.5ex \noindent
One typically asks for a high probability bound on the above error in terms of $n$.
Such error bounds have been established for a variety of 
partitioning algorithms including aforementioned approaches. \citet{Chen_2014_jour_TIT} 
compare the theoretical guarantees for different approaches.

In the case of random graphs under stochastic block model, analysis
of a spectral algorithm was first considered by~\citet{McSherry_2001_conf_FOCS}.
However, the popular variant of spectral graph partitioning, commonly known as
\emph{spectral clustering}, was studied only in recent 
times~\citep{Rohe_2011_jour_AnnStat,Lei_2015_jour_AnnStat}.
It is now known that for a planted graph with $\Omega(\ln n)$ minimum 
node degree, spectral clustering achieves an $o(n)$ error rate.
One commonly refers to this property as the \emph{weak consistency} of spectral clustering.
This not the best known error rate as exact recovery of the partitions are known to be
possible using other approaches~\citep{Amini_2014_arxiv}. Recent results~\citep{Vu_2014_arxiv,Lei_2014_arxiv,Gao_2015_arxiv} show that an additional refinement
process can improve the partitioning of spectral clustering to exactly recover the partitions,
thereby achieving \emph{strong consistency}.
The condition on minimum node degree can be also relaxed by considering alternative
spectral techniques~\citep{Krzakala_2013_jour_PNAS,Le_2015_arxiv}, and these algorithms
can detect partitions in sparse random graphs that are close to the algorithmic barrier
for community detection~\citep{Decelle_2011_jour_PhyRevE}.

\subsection*{Hypergraph partitioning}
In spite of the vast applicability of network modeling and analysis, there exists more
complex scenarios, where pairwise interactions cannot accurately model the system
of interest. A common example is folksonomy, where individuals annotate online 
resources, such as images or research papers. 
Such problems appear to have a tri-partite structure in form of 
``user--resource--annotation'', and is
naturally represented as a 3-uniform hypergraph \citep{Ghoshal_2009_jour_PhyRevE},
where each edge connects three nodes.
Earlier works in data mining~\citep{Gibson_2000_jour_VLDB} 
as well as in computer vision~\citep{Govindu_2005_conf_CVPR} have also 
demonstrated the necessity of uniform hypergraphs.
Moreover, in large scale circuit design~\citep{Karypis_2000_jour_VLSI}
and molecular interaction networks~\citep{Michoel_2012_jour_PhyRevE}, 
one needs to consider group interactions that is appropriately modeled by a
non-uniform hypergraph.

The current work focuses on hypergraph partitioning that appears in various applications such 
as circuit partitioning~\citep{Schweikert_1979_conf_DesignAutomation},
categorical data clustering~\citep{Gibson_2000_jour_VLDB},
geometric grouping~\citep{Govindu_2005_conf_CVPR} and others.
Various partitioning techniques are used in practice including move-based algorithms~\citep{Schweikert_1979_conf_DesignAutomation,Karypis_2000_jour_VLSI},
spectral algorithms \citep{Rodriguez_2002_jour_LinMultilinAlg,Zhou_2007_conf_NIPS}, 
tensor based methods~\citep{Govindu_2005_conf_CVPR} etc.

Hypergraph partitioning and related problems have been of theoretical interest for
quite some time~\citep{Berge_1984_book_Elsevier}. 
While early works on hypergraph partitioning studied various properties of hypergraph 
cuts~\citep{Bolla_1993_jour_DiscreteMath,Chung_1992_conf_DIMACS}, more recent results
provide insights into the algebraic connectivity and chromatic numbers of 
hypergraphs~\citep{Hu_2012_jour_JCombOptim,Cooper_2012_jour_LinearAlgebra}.
However, till date, little is known about the theoretical guarantees of hypergraph partitioning 
methods that are popular amongst practitioners.
The primary reason for this lack of results, at least in a stochastic framework,
is due to the absence of random models of non-uniform hypergraphs 
that can incorporate a planted structure.
Planted structures in uniform hypergraphs have been 
studied in context of hypergraph coloring~\citep{Chen_1996_jour_IPCO}, 
and in a more general setting in~\citep{Ghoshdastidar_2014_conf_NIPS},
where almost sure error bounds are derived for partitioning planted uniform hypergraphs
via tensor decomposition. But, extension of such models or analysis to non-uniform 
hypergraphs does not follow directly.

\subsection*{Contributions}
The primary focus of this paper is to derive an error bound for a
hypergraph partitioning algorithm that solves a spectral 
relaxation of the normalized hypergraph cut problem. 
This is achieved in the form of a two-fold contribution. 

We present a model for generating random hypergraphs with a planted solution.
Extensions of the Erd\"os-R\'enyi model to non-uniform hypergraphs have been 
studied in the literature~\citep{SchmidtPruzan_1985_jour_Combinatorica,Darling_2005_jour_AnnApplProbab},
where, for each $m$, the probability of generating edges of size $m$ is controlled
by a parameter $p_m$. 
The recent work of~\citet{Stasi_2014_arxiv} present a similar model, but with a specified 
degree sequence.
Such models implicitly suggest that one can consider a non-uniform
hypergraph as a collection of $m$-uniform hypergraphs for varying $m$. 
Thus, it is possible to construct planted models for non-uniform hypergraphs from a
collection of uniform hypergraph models.
Based on this idea, we present a planted hypergraph model that
naturally extends the sparse stochastic block model for graphs~\citep{Lei_2015_jour_AnnStat},
and also encompasses previously studied models for uniform 
hypergraphs~\citep{Ghoshdastidar_2014_conf_NIPS}.

We consider a popular spectral algorithm for hypergraph partitioning, and
derive a bound on the number of nodes incorrectly assigned by the algorithm under the above model.
We prove that for random planted hypergraphs with minimum node degree above a certain
threshold, the algorithm is weakly consistent in general.
However, the algorithm can also exactly recover the partitions from dense hypergraphs
without any subsequent refinement procedure.
Our analysis relies on an alternative characterization of the incidence matrix of the 
random hypergraph, and the use of matrix concentration 
inequalities~\citep{Chung_2011_jour_EJComb,Tropp_2012_jour_FOCM}.

Typically, spectral partitioning algorithms involve a post-processing stage of
distance based clustering.
Though the $k$-means algorithm~\citep{Llyod_1982_jour_TIT} or its approximate variants~\citep{Kumar_2004_conf_FOCS,Ostrovsky_2012_jour_JACM} are the popular choice in practice, 
such algorithms are not always guaranteed to provide good clustering. 
\citet{Gao_2015_arxiv} discusses the implication of this  drawback on the consistency results
for spectral clustering~\citep{Lei_2015_jour_AnnStat}. On the other hand,
we establish that under certain conditions, 
the approximate $k$-means algorithm  of~\citep{Ostrovsky_2012_jour_JACM}
indeed provides a good clustering with very high probability.

Finally, we consider special cases of the planted model.
We comment on the allowable model parameters, and illustrate their effect on the 
derived error bound.
Numerical studies 
reveal the practical
significance of spectral hypergraph partitioning as well as the applicability of our analysis.

\subsection*{Organization}
We first describe the 
spectral hypergraph partitioning algorithm under consideration in Section~\ref{sec_algorithm}. 
Section~\ref{sec_model} describes the model for random hypergraphs with a planted partition.
We provide the main consistency result in Section~\ref{sec_main_result},
followed by a series of examples of planted models studied in Section~\ref{sec_specialcases}.
Section~\ref{sec_expt} contains experimental results that 
validate our model and analysis, and Section~\ref{sec_conclusion} presents
the concluding remarks.
The proofs of the technical lemmas can be found in the appendices.

\subsection*{Notations}
Some of the notations that is often used in this paper are mentioned here.
$\1\{\cdot\}$ is the indicator function and
$\ln(\cdot)$ refers to the natural logarithm.
$\E[\cdot]$ denotes expectation with respect to the distribution of the planted model.
For a matrix $A$, we use $A_{i\bull}$ to refer to the $i^{th}$ row of $A$
and $A_{\bull i}$ refers to its $i^{th}$ column.
$\Vert\cdot\Vert_2$ denotes the Euclidean norm for vectors and the spectral norm for
matrices, while $\Vert\cdot\Vert_F$ denotes the Frobenius norm.
We sometimes compute standard matrix functions like Trace$(\cdot)$ and det$(\cdot)$.
In addition, we also use asymptotic notations $O(\cdot), o(\cdot), \Omega(\cdot)$ etc.,
where we view these quantitites as functions of  the number of nodes $n$.

\section{Spectral Hypergraph Partitioning}
\label{sec_algorithm}
A hypergraph is defined as a tuple $(\Vc,\Ec)$, where $\Vc$ is a set of objects
and $\Ec$ is a collection of subsets of $\Vc$.
Though early works in combinatorics viewed this structure purely as a set 
system, it was soon realized that one may view
$\Vc$ as a set of nodes and every element of $\Ec$ as an edge (or connection)
among a subset of nodes.
As noted in~\citep{Berge_1984_book_Elsevier}, such a generalization of graphs
helps to simplify several combinatorial results in the graph literature.
A hypergraph is said to be
$r$-uniform if every edge $e\in\Ec$ contains exactly $r$ nodes. 

In this paper, we assume that there are no edges of size 0 or 1 as they do not convey
any information in a partitioning framework.  
%
We also assume that the hypergraph is undirected, \ie there is no ordering of nodes
in any edge. 
Under this setting, the most simple representation of a hypergraph is in terms of its
incidence matrix $H\in\{0,1\}^{|\Vc|\times|\Ec|}$, where $H_{ve} = 1$ if
the node $v$ is contained in the edge $e$, and 0 otherwise.
One can note that the degree of any node $v$ can be written as
$\degr(v) =  \sum_{e\in\Ec} H_{ve}$, which is simply the sum of 
the $v^{th}$ row of $H$. Similarly, the cardinality of any edge $e$ is
$|e| = \sum_{v\in\Vc} H_{ve}$.

Several notions of hypergraph cut and hypergraph Laplacian have been proposed
in the literature~\citep{Bolla_1993_jour_DiscreteMath,Chung_1992_conf_DIMACS,Rodriguez_2002_jour_LinMultilinAlg}
that generalize the standard notions well studied in the graph literature.
In this work, we consider the generalization studied in~\citep{Zhou_2007_conf_NIPS}.
Let $\Vc_1\subset\Vc$, then
$\vol(\Vc_1) = \sum_{v\in\Vc_1}\degr(v)$ is called the volume of $\Vc_1$,
while the boundary of $\Vc_1$, 
defined as $\partial\Vc_1 = \{e\in\Ec : e\cap\Vc_1 \neq \phi, e\cap\Vc_1^c\neq\phi\}$,
denotes the set of edges that are cut when the nodes are divided into $\Vc_1$ and 
$\Vc\backslash\Vc_1$. 
The volume of $\partial\Vc_1$ is defined as
\begin{align*}
\vol(\partial\Vc_1) = \sum_{e\in\partial\Vc_1} \frac{|e\cap\Vc_1||e\cap\Vc_1^c|}{|e|}\;.
\end{align*}
We consider the problem of partitioning the vertex set $\Vc$ into $k$ 
disjoint sets, $\Vc_1,\ldots,\Vc_k$, that minimizes the normalized hypergraph cut 
\begin{align}
\textup{NH-cut}(\Vc_1,\ldots,\Vc_k) = \sum_{j=1}^k \frac{\vol(\partial\Vc_j)}{\vol(\Vc_j)} \;.
\label{eq_nhcut_defn}
\end{align}
One can observe that for graphs, the above definition~\eqref{eq_nhcut_defn}
retrieves the standard notion of a normalized cut~\citep{vonLuxburg_2007_jour_StatComp}.
\citet{Zhou_2007_conf_NIPS} also define the notion of a normalized hypergraph Laplacian
matrix $L\in\R^{|\Vc|\times|\Vc|}$ given by
\begin{align}
 L = I - D^{-1/2}H\Delta^{-1}H^TD^{-1/2},
 \label{eq_hyperlap}
\end{align}
where the matrices $D\in\R^{|\Vc|\times|\Vc|}, \Delta\in\R^{|\Ec|\times|\Ec|}$ 
are diagonal with $D_{vv} = \degr(v)$ and $\Delta_{ee} = |e|$.
A simple calculation shows that
the problem of minimizing the quantity in~\eqref{eq_nhcut_defn} is equivalent to 
the problem:
\begin{align}
 \underset{\Vc_1,\ldots,\Vc_k}{\textup{minimize}} \textup{~~Trace}\left(\widehat{X}^T
 L\widehat{X} \right),
 \label{eq_nhcut_min}
\end{align}
where $\widehat{X}\in\R^{|\Vc|\times k}$ is such that $\widehat{X}_{vj} = \sqrt{\frac{\deg(v)}{\vol(\Vc_j)}}\1\{v\in\Vc_j\}$, 
and satisfies $\widehat{X}^T\widehat{X} = I$.
Since the optimization in~\eqref{eq_nhcut_min} is NP-hard,
one relaxes the problem by minimizing over all ${X}\in\R^{|\Vc|\times k}$ with orthonormal columns. It is well known that
the solution to this relaxed problem is the matrix of $k$ leading orthonormal 
eigenvectors of $L$.
Note that $L$ is a positive semi-definite matrix
with at least one eigenvalue equal to zero. 
The term ``leading eigenvectors'' refers to the eigenvectors that
correspond to the $k$ smallest eigenvalues of $L$.

The above discussion motivates a spectral $k$-way partitioning approach based on minimizing NH-cut. The method is listed in Algorithm~\ref{alg_SHP}.
The form of Laplacian matrix in~\eqref{eq_hyperlap} also suggests that the problem
of minimizing NH-cut may be alternatively expressed as the problem of partitioning a graph
with weighted adjacency matrix
\begin{align}
 A = H\Delta^{-1}H^T\;.
\label{eq_A_defn}
\end{align}
Such a graph is related to
the star expansion of the hypergraph~\citep{Agarwal_2006_conf_ICML}.

The intuition behind the $k$-means step in Algorithm~\ref{alg_SHP} is as follows. 
If the solution of the spectral relaxation results in $X=\widehat{X}$,
where $\widehat{X}$ is defined as in~\eqref{eq_nhcut_min},
then after row normalization, $\overline{X}$ corresponds to a binary matrix with exactly 
one non-zero term in each row.
Hence, one obtains the partitions desired in~\eqref{eq_nhcut_min}
by performing $k$-means on the rows of $\overline{X}$.
In this paper, we assume that the approximate $k$-means 
method of~\citep{Ostrovsky_2012_jour_JACM} is used, that provides a near optimal solution 
in a single iteration.
 
\begin{algorithm}[h]
\caption{Spectral Hypergraph Partitioning Algorithm}
 \begin{algorithmic}
 \INPUT Incidence matrix $H$ of the hypergraph.\\
 \STATE 1: Compute the hypergraph Laplacian $L$ as in~\eqref{eq_hyperlap}.\\
 \STATE 2: Compute the leading eigenvector matrix $X\in\R^{|\Vc|\times k}$.\\
 \STATE 3: Normalize rows of $X$ to have unit norm. Call this matrix $\overline{X}$.\\
 \STATE 4: Run $k$-means on the rows of $\overline{X}$.\\
 \OUTPUT Partition of $\Vc$ that corresponds to the clusters obtained from $k$-means.\\
 \end{algorithmic}
 \label{alg_SHP}
\end{algorithm}
 
\section{Planted Partition in Random Hypergraphs}
\label{sec_model}

In the rest of the paper, we study the 
error incurred by Algorithm~\ref{alg_SHP}. 
For this, we consider a model for generating random hypergraphs 
with a planted solution.

\subsection{The model}
Let $\Vc = \{1,2,\ldots n\}$ be a set of nodes, and let
$\psi:\{1,2,\ldots,n\} \to\{1,2,\ldots,k\}$  be a partition of the
nodes into $k$ classes. Here, $\psi$ is the unknown planted
partition that one needs to extract from a hypergraph generated on
$\Vc$. For a node $i$, we denote its class by $\psi_i$. 
%
Let $M\geq2$ be an integer, representing the range or the maximum edge cardinality in the hypergraph.
In view of practical situations, we allow both $k$ and $M$ to vary with $n$ though
this dependence is not made explicit in the notation.
One may set $M = n$ to allow occurrence of all possible edges, but in 
practice, one can assume that $M=O(\ln n)$.
Also, for each $n$ and for each $m=2,\ldots,M$, 
let $\alpha_{m,n}\in[0,1]$, and $B^{(m)} \in [0,1]^{k \times k \times \ldots \times k}$
be a symmetric $k$-dimensional tensor of order $m$.

A random hypergraph on $\Vc$ is generated as follows. For each $m=2,\ldots,M$,
and for every set $\{i_1,i_2,\ldots,i_m\}\subset\Vc$, 
an edge is included independently with probability 
$\alpha_{m,n}B^{(m)}_{\psi_{i_1}\psi_{i_2}\ldots\psi_{i_m}}$.
This process generates a random hypergraph of maximum edge cardinality $M$.
The tensor $B^{(m)}$ contains the probabilities of forming $m$-way edges
among the different classes if $\alpha_{m,n}=1$.
On the other hand, $\alpha_{m,n}$ allows for a 
sparsity scaling that does not depend on the partitions.
In the case of sparse graphs, $\alpha_{2,n}$ regulates the edge density.
However, in real-world non-uniform hypergraphs, one often finds than the density of 2 or
3-way edges is much more than edges of larger size (say, 10). To account for 
this generality, we allow $\alpha_{m,n}$ to vary both with $m$ and $n$.
For instance, if $\alpha_{2,n}=1$ and $\alpha_{m,n} =\frac{1}{n^{m-1}}$ for all $m>2$, then 
the generated hypergraph contains $O(n^2)$ number of 2-way edges, but only $O(n)$ 
number of $m$-way edges for every $m>2$.

As a special case, note that for graphs, $M=2$ for all $n$, and 
the model corresponds to the sparse stochastic block model,
where an edge $(u,v)$ is formed with probability $\alpha_{2,n}B^{(2)}_{\psi_u\psi_v}$.
In other words, if $Z\in\{0,1\}^{n\times k}$ denotes the assignment matrix, then 
the probability of edge $(u,v)$ is 
same as the corresponding entry of $\alpha_{2,n}ZB^{(2)}Z^T$.
For $r$-uniform uniform hypergraphs, one has $\alpha_{m,n} = 0$
for all $m\neq r$. \cite{Ghoshdastidar_2014_conf_NIPS} considered a dense
uniform hypergraph, \ie $\alpha_{r,n} = 1$, and edge probabilities specified by a
$r^{th}$-order $k$-dimensional tensor $B^{(r)}$.
It was shown that the population adjacency tensor can be expressed in terms of $B^{(r)}$ and $Z$.

The intuition behind the above described model is that one may view a hypergraph
of range $M$ as a collection of uniform hypergraphs of orders 
$m=2,\ldots,M$. In the random setting, each $m$-uniform hypergraph is specified
in terms of $\alpha_{m,n}$ and $B^{(m)}$.
The above model can be easily extended to directed hypergraphs,
and also to the case of  weighted hypergraphs.

\subsection{The random hypergraph Laplacian}
For stochastic block model, a random instance of a graph is specified by its $n\times n$ adjacency matrix.
However, for random hypergraphs, the size of the incidence matrix $H$
is a random quantity as it depends on the number of generated
edges. Since this poses difficulties in working with the form of
hypergraph Laplacian in~\eqref{eq_hyperlap}, we present an alternative
representation. The Laplacian can be written as
\begin{align}
 L = I - \sum_{e\in\Ec} \frac{1}{|e|} D^{-1/2}a_e a_e^TD^{-1/2} \enspace,
 \label{eq_hyperlap1}
\end{align}
where for $e\subset\Ec$, $a_e \in \{0,1\}^n$ with
$(a_e)_i = 1$, if node $i\in e$, and 0 otherwise.

Let $\beta_M = \displaystyle\sum_{m=2}^M \binom{n}{m}$. Note that 
$\beta_M$ is the maximum number of edges the hypergraph can contain given 
the fact that its range is $M$.
For convenience, we define a bijective map 
$\xi:\{1,2,\ldots,\beta_M\}\to\{e\subset \Vc: 2\leq |e| \leq M\}$,
where each $\xi_j$ refers to a subset of nodes, \ie a possible
edge in the given hypergraph.
Then the Laplacian can be expressed as
\begin{align}
 L = I - \sum_{j=1}^{\beta_M} \frac{\1\{\xi_j\in{\Ec}\}}{|\xi_j|} D^{-1/2}a_{\xi_j} a_{\xi_j}^TD^{-1/2},
 \label{eq_hyperlap2}
\end{align}
where the summation is over all possible edges of size at most $M$, but the 
missing edges do not contribute to the sum. Similarly, one can express the degree matrix $D$ as
\begin{align}
 D_{ii} = \degr(i) 
 = \sum_{e\in\Ec} (a_e)_i = \sum_{j=1}^{\beta_M} \1\{\xi_j\in{\Ec}\}(a_{\xi_j})_i.
 \label{eq_degree1}
\end{align}
%
%
The above representation corresponds to an `extended'
version of the incidence matrix as $\overline{H}\in\{0,1\}^{n\times \beta_M}$,
whose $j^{th}$ column is $\1\{\xi_j\in{\Ec}\}a_{\xi_j}$,
\ie $\overline{H}$ contains the columns of $H$ with
additional zero columns inserted to account for missing edges. 
This holds for any hypergraph of range $M$ defined on the set $\Vc$.
We use this representation to keep the number of columns as a deterministic quantity.
We now discuss how the described planted partition model for hypergraphs, with
maximum edge size $M$, can be expressed in terms of 
the extended incidence matrix $\overline{H} \in\{0,1\}^{n \times \beta_{M}}$.
Let $h_j$, $j=1,2,\ldots,\beta_{M}$ be independent Bernoulli random variables
that indicate the presence of the edge $\xi_j \subset\Vc$.
By description of the model, if $\xi_j = \{i_1,i_2,\ldots,i_{m_j}\}$ for some 
$m_j\in\{2,\ldots,M\}$, then the random variable $h_j \sim$
Bernoulli$\left(\alpha_{m_j,n}B^{(m_j)}_{\psi_{i_1}\psi_{i_2}\ldots\psi_{i_{m_j}}}\right)$.
The $j^{th}$ column of $\overline{H}$ is $h_j a_{\xi_j}$,
and hence, the Laplacian matrix for the random hypergraph is
\begin{align}
 L = I - \sum_{j=1}^{\beta_{M}} \frac{h_j}{|\xi_j|} D^{-1/2}a_{\xi_j} a_{\xi_j}^TD^{-1/2},
 \qquad \text{where }
 D_{ii} = \sum_{j=1}^{\beta_{M}} h_j(a_{\xi_j})_i.
 \label{eq_hyperlap_random}
\end{align}
At this stage, we note that the above matrices depend on the number of nodes $n$.
For ease of notation, we do not explicitly mention this dependence.

\section{Consistency of Spectral Hypergraph Partitioning}
\label{sec_main_result}

This section presents the main result of this paper that gives
a bound on the error incurred by the 
spectral hypergraph partitioning algorithm described in Algorithm~\ref{alg_SHP}. 
Let $\psi':\{1,\ldots,n\}\to\{1,\ldots,k\}$ denote the
labels obtained from the algorithm. The partitioning error is given by
the number of nodes incorrectly assigned  by Algorithm~\ref{alg_SHP}, \ie
\begin{equation}
 \err = \min_\sigma \sum_{i=1}^n \1\{\psi_i\neq\sigma(\psi_i')\}\;,
\end{equation}
where the minimum is taken over all permutation $\sigma$ of labels.
We show that if {\bf(i)} the partitions are \emph{identifiable}, and {\bf(ii)} 
the hypergraph is not too \emph{sparse}, then indeed $\err$ is bounded by a quantity that is at 
most sub-linear in $n$. 
Furthermore, the bound holds with probability $(1-o(1))$.
This immediately implies that Algorithm~\ref{alg_SHP} is \emph{weakly consistent}.
However, we show later that for particular model parameters, Algorithm~\ref{alg_SHP} can 
even recover the partitions exactly, \ie $\err=o(1)$.

\subsection{The main result}
The consistency result studied in this paper is quite similar, in spirit, to those studied in the 
case of stochastic block model for graphs.
In such a case, one typically analyzes the population version of a spectral algorithm,
and then uses the fact that the spectral properties of the Laplacian eventually concentrates 
around those of the population Laplacian.
 
From this point of view, we consider the population version of the hypergraph 
Laplacian~\eqref{eq_hyperlap_random} defined as
\begin{align}
 \Lc = I - \sum_{j=1}^{\beta_{M}} \frac{\E[h_j]}{|\xi_j|} \Dc^{-1/2}a_{\xi_j} a_{\xi_j}^T\Dc^{-1/2}\;,
 \label{eq_hyperlap_population}
\end{align}
where $\Dc$ is the expected degree matrix, \ie 
$\Dc_{ii} = \sum\limits_{j=1}^{\beta_{M}}\E[h_j](a_{\xi_j})_i$.
We also define the quantity $d = \min\limits_{i\in\{1,\ldots,n\}} \Dc_{ii}$.
Without loss of generality, we may also assume that for a given $n$,
the community sizes are  $n_1 \geq n_2 \geq \ldots \geq n_k$ 

Before stating the main result, it is useful to elaborate on the aforementioned conditions
under which the derived error bound holds.
A lower bound on the sparsity of the hypergraph is a standard requirement
to ensure that the concentration of the spectral properties eventually hold,
and has been often used in the graph literature~\citep{Lei_2015_jour_AnnStat,Le_2015_arxiv}.
In our setting, this can be stated in terms of the sparsity factors $\alpha_{m,n}$,
or more simply, in terms of the minimum expected degree  $d$,
that grows with $n$ but at a rate controlled by the sparsity factors.

A more critical condition is the identifiability of the partitions.
Note that the definition of the hypergraph Laplacian essentially implies that 
the hypergraph is reduced to a graph with self loops. 
Hence, the performance of Algorithm~\ref{alg_SHP} crucially depends on the identifiability
of the partitions from $\Lc$, or rather from this reduced graph
with the population adjacency matrix 
\begin{align*}
 \Ac = \E[A] = \sum_{j=1}^{\beta_{M}} \frac{\E[h_j]}{|\xi_j|} a_{\xi_j} a_{\xi_j}^T\;.
\end{align*}
The following result provides a characterization of $\Lc$ and $\Ac$,
which in turn helps to quantify the condition for identifiability of the partitions from $\Lc$.
\begin{lemma}
\label{lem_expected1}
Let $Z\in\{0,1\}^{n\times k}$ denote the assignment matrix corresponding to the partition $\psi$.
Then the population hypergraph Laplacian is given by
\begin{align}
 \Lc = I - \Dc^{-1/2}\Ac\Dc^{-1/2}\;,
\end{align}
where $\Ac$ can be expressed as
\begin{align}
 \Ac = ZGZ^T - J \;.
 \label{eq_Ac_matrix_form}
\end{align} 
Here, $J\in\R^{n\times n}$ is diagonal with $J_{ii} = J_{jj}$ whenever $\psi_i=\psi_j$,
and $G\in\R^{k\times k}$.

Furthermore, $\Lc$ contains $k$ eigenvalues for which the corresponding orthonormal
eigenvectors are the columns of the matrix $\Xc = Z(Z^TZ)^{-1/2}U$,
where $U\in\R^{k\times k}$ is orthonormal.
\end{lemma}
The representation in~\eqref{eq_Ac_matrix_form} shows that $\Ac$ is essentially of rank $k$,
except for the diagonal entries. Owing to the first term in~\eqref{eq_Ac_matrix_form}, 
one does expect $\Lc$ to have $k$ eigenvectors
whose entries are constant in each community. As discussed later, a close inspection of $\Xc$
 reveals that indeed 
the columns of $\Xc$ satisfy this property. Thus, if the spectral stage of Algorithm~\ref{alg_SHP} can extract $\Xc$,
then zero error can be achieved from the $k$-means step.

In general, $\Xc$ need not correspond to leading eigenvectors $\Lc$ (as computed in Algorithm~\ref{alg_SHP}).
This is true even for certain types of graphs, for instance $k$-colorable graphs~\citep{Alon_1997_jour_SIAMJComput}.
This effect is more pronounced in non-uniform hypergraphs due to the presence of a large 
number of model parameters.
To account for this factor, we define the following quantity
\begin{align}
\delta 
&= \left(\lambda_{\min}(G)\min_{1\leq i\leq n} 
\frac{n_{\psi_i}}{\Dc_{ii}}\right) - 
\max_{1\leq i,j\leq n} \left|\frac{J_{ii}}{\Dc_{ii}} 
- \frac{J_{jj}}{\Dc_{jj}}\right|\;,
\label{eq_delta_charac}
\end{align}
where $n_{\psi_i}$ is the size of the community in which node $i$ belongs.
We show that if $\delta>0$, then the columns of $\Xc$ are the $k$ leading eigenvectors 
of $\Lc$. Here, $\lambda_{\min}(G)$ refers to the smallest eigenvalue of $G$.
Thus, we can state the consistency result for Algorithm~\ref{alg_SHP} as below.
\begin{theorem}
\label{thm_main_consistency}
Consider a random hypergraph on $n$ nodes generated according to the
planted partition model described in Section~\ref{sec_model}.
Assume that $n$ is sufficiently large, and
the size of the $k$ partitions are $n_1\geq n_2 \geq \ldots \geq n_k$.
Let $d$ be the minimum expected degree, and $\delta$ be the quantity
defined in~\eqref{eq_delta_charac}. 

There exists an absolute constant $C>0$, such that, if 
$\delta>0$ and 
\begin{align}
d > C\frac{kn_1 (\ln n)^2}{\delta^2 n_k}
\label{eq_degree_condn}
\end{align}
then with probability at least $1-O\left((\ln n)^{-1/4}\right)$,
\begin{equation}
\err = O\left( \frac{kn_1 \ln n}{\delta^2 d}\right).
\label{eq_error_bound}
\end{equation}
\end{theorem}
Note here that the quantities $\delta,d$ and $k$ can vary with $n$.
On substituting the condition on $d$ into~\eqref{eq_degree_condn}, one can see that
$\err = o(n)$ with probability $(1-o(1))$. Hence, Algorithm~\ref{alg_SHP}
is weakly consistent if the conditions of the theorem are satisfied.
However, we show later that in certain dense hypergraphs, the bound in~\eqref{eq_error_bound} may eventually decay to zero.
Thus, Algorithm~\ref{alg_SHP} is guaranteed to exactly recover the communities in such cases.

In Section~\ref{sec_specialcases}, we consider particular instances  of the planted model,
and illustrate the dependance of the above result on the model parameters.
For instance, \eqref{eq_degree_condn} implies that the result holds if the sparsity factor 
$(\alpha_{m,n})$ is above a certain threshold
(see Corollaries~\ref{cor_equal_size_uniform} and~\ref{cor_equal_size_nonuniform}). 
Even when \eqref{eq_degree_condn} holds, higher error is incurred for a sparse hypergraph 
(small $d$) or when the number of communities $k$ is large.

One may note that $\delta>0$ is the condition for identifiability of the partitions,
and is essential for success of the algorithm. 
Typically, one does find that $\delta\downarrow0$
as $n\to\infty$. To this end, the condition~\eqref{eq_degree_condn} implies that $\delta$
cannot decay rapidly as $\delta^2d$ needs to maintain a minimum growth rate.
We also note that $\delta$ quantifies identifiability of the partitions and $\err$ varies as 
$\frac{1}{\delta^2}$. Hence, if the model parameters
are such that $\delta$ is small, for instance if the
probability of inter-community edges is very close
to that of within community edges, then $\err$ is larger.

Before presenting the proof of Theorem~\ref{thm_main_consistency}, we comment on the assumption of sufficiently large $n$.
Note that the sole purpose of this assumption is to ensure the success  
of the $k$-means algorithm. Later, in the proof, we establish that if $n$ is large enough,
the condition~\eqref{eq_degree_condn} ensures that the approximate $k$-means
method of~\citet{Ostrovsky_2012_jour_JACM} provides a near optimal solution, 
which is worse by only a constant factor.
Earlier works on spectral graph partitioning~\citep{Rohe_2011_jour_AnnStat,Lei_2015_jour_AnnStat}
assumed the existence of such a near optimal solution with probability 1.
To demonstrate the effect of such an assumption,
we state the following result, 
which is a modification of Theorem~\ref{thm_main_consistency} under the above assumption.
\begin{corollary}
\label{cor_prob1_kmeans}
Consider a random hypergraph on $n$ nodes generated according to the
planted partition model, and 
let the other quantities be as defined in Theorem~\ref{thm_main_consistency}.
Assume that for a constant $\gamma>1$,
there is a $\gamma$-approximate\footnote{
Informally, a $\gamma$-approximate $k$-means methods
returns a solution for which the objective of the $k$-means problem is at most
$\gamma$ times the global minimum, where $\gamma>1$.
A formal definition is postponed to~\eqref{eq_suboptimal_kmeans} and subsequent discussions. 
}
 $k$-means algorithm that succeeds with probability 1.

There exists an absolute constant $C>0$, such that, if 
$\delta>0$ and 
\begin{align}
d > C\frac{\ln n}{\delta^2}
\label{eq_degree_condn1}
\end{align}
then with probability at least $1-\frac{4}{n^2}$,
\begin{equation}
\err = O\left( \frac{kn_1 \ln n}{\delta^2 d}\right).
\label{eq_error_bound1}
\end{equation}
\end{corollary}

The result reveals that if a good $k$-means algorithm is available, then the success
probability of Algorithm~\ref{alg_SHP} increases, and the result is also applicable
for more sparse hypergraph since the condition~\eqref{eq_degree_condn1}
is weaker than~\eqref{eq_degree_condn}.
However, based on the existing results in the $k$-means literature, 
one should consider the following remark.
\begin{remark}
If the data satisfies certain \emph{clusterability} criterion\footnote{Various clusterability criteria have been
studied in the literature. In this work, we consider the notion of $\epsilon$-separability
proposed by~\citet{Ostrovsky_2012_jour_JACM}.},
then the efficient variants of $k$-means~\citep{Kumar_2004_conf_FOCS,Ostrovsky_2012_jour_JACM}
provide a $\gamma$-approximate solution with a constant probability $\rho<1$.
Both $\gamma$ and $\rho$ depend on various factors including $k$, clusterability criterion etc.
\end{remark}

In view of the above remark, Corollary~\ref{cor_prob1_kmeans} is too optimistic.
Recently, \citet{Gao_2015_arxiv} pointed that if one uses the method of
\citet{Kumar_2004_conf_FOCS}, then $\gamma$ grows with $k$. In addition, one should
also note that the success probability of this method is $\rho=c^k$ for an absolute constant $c\in(0,1)$. Hence, a spectral partitioning algorithm using this method cannot succeed
with probability $(1-o(1))$. Instead, we use the method of~\citet{Ostrovsky_2012_jour_JACM}
to achieve a higher success rate as stated in Theorem~\ref{thm_main_consistency}.
The only additional assumption is that of sufficiently large $n$.
We note that this requirement, along with condition~\eqref{eq_degree_condn},
can be relaxed if one only aims for a constant success probability.
This is shown in the following modification of Theorem~\ref{thm_main_consistency},
where we assume that the $k$-means algorithm of~\citet{Ostrovsky_2012_jour_JACM} is used.

\begin{corollary}
\label{cor_constprob_kmeans}
Consider a random hypergraph on $n$ nodes generated according to the
planted partition model, and 
let the other quantities be as defined in Theorem~\ref{thm_main_consistency}.

There exist absolute constants $C>0$ and $\epsilon\in(0,0.015)$, such that, if 
$\delta>0$ and 
\begin{align}
d > \frac{C}{\epsilon^2}\frac{k n_1\ln n}{\delta^2 n_k}
\label{eq_degree_condn2}
\end{align}
then with probability at least $1-O(\sqrt{\epsilon})$,
\begin{equation}
\err = O\left( \frac{kn_1 \ln n}{\delta^2 d}\right).
\label{eq_error_bound2}
\end{equation}
\end{corollary}

\subsection{Proof of Theorem~\ref{thm_main_consistency}}

We now present an outline of the proof of Theorem~\ref{thm_main_consistency}
using a series of lemmas. The proofs for these 
lemmas are given in the appendix.
The result is obtained by proving the following facts:
\begin{enumerate}
\item
If Algorithm~\ref{alg_SHP} is performed on the population Laplacian $\Lc$, then
under the condition of $\delta>0$, the obtained partitions are correct.
\item
The deviation of $L$ from $\Lc$ is bounded above, and the bound holds with probability
at least $(1-\frac{4}{n^2})$.
\item
As a consequence of above facts, the standard matrix perturbation
bounds~\citep{Stewart_1990_book_AP} imply that the eigenvalues and 
the corresponding eigenspaces of $L$ concentrate about those of $\Lc$.
\item
If~\eqref{eq_degree_condn} holds, then $k$-means stage of Algorithm~\ref{alg_SHP}
succeeds in obtaining a near optimal solution with probability at least 
$1-O\left((\ln n)^{-1/4}\right)$.
\item
The partitioning error can be expressed in terms of the above bounds, which leads 
to~\eqref{eq_error_bound}. 
\end{enumerate}
Corollaries~\ref{cor_prob1_kmeans} and~\ref{cor_constprob_kmeans} can be proved
in similar manner. This is discussed in the appendix.
We now prove the above facts.
The following result extends
Lemma~\ref{lem_expected1}.
\begin{lemma}
\label{lem_expected}
If $\delta>0$,
then the $k$ leading orthonormal eigenvectors of $\Lc$ correspond to the columns of the 
matrix  $\Xc = Z(Z^TZ)^{-1/2}U$.
\end{lemma}
In the above result, $Z^TZ$ is a diagonal matrix with entries being
the sizes of the $k$ partitions. Hence, both $Z^TZ$ and $U$ are of
the rank $k$. 
Due to this, one can observe that the matrix $\Xc$ contains exactly $k$ distinct rows, each corresponding to a particular partition, \ie
if $A_{i\bull}$ denotes $i^{th}$ row of a matrix $A$, then for any two nodes $i,j\in\Vc$,
\begin{align*}
\Xc_{i\bull} = \Xc_{j\bull}
\Longleftrightarrow Z_{i\bull} = Z_{j\bull}
\Longleftrightarrow \psi_i = \psi_j \;.
\end{align*}
Moreover, since $U$ is orthonormal, the distinct rows of $\Xc$ are orthogonal. 
Hence, after row normalization, the distinct rows correspond to $k$ orthonormal
vectors in $\R^{k}$, which can be easily clustered by $k$-means algorithm
to obtain the true communities.
%
Technically, $\delta$ is a lower bound on
the eigen-gap between the $k^{th}$ and $(k+1)^{th}$ smallest eigenvalues of $\Lc$. 
Since, it is difficult to obtain a simple characterization of the eigen gap,
we resort to the use of $\delta$ as defined in~\eqref{eq_delta_charac}.

Next, we bound the deviation of a random instance of $L$ from the population 
Laplacian $\Lc$. 
This bound relies on the use of matrix Bernstein
inequality~\citep{Chung_2011_jour_EJComb,Tropp_2012_jour_FOCM}.
We note that for graphs, sharp deviation bounds have been
used~\citep{Lei_2015_jour_AnnStat}, but such techniques cannot be directly 
extended to the case of hypergraphs.
\begin{lemma}
\label{lem_random_deviation}
If $d>9\ln n$, then
with probability at least $(1-\frac{4}{n^2})$,
\begin{equation}
\Vert L - \Lc \Vert_2 \leq 12\sqrt{\frac{\ln n}{d}} \;.
\end{equation}
\end{lemma}

We now use the principle subspace perturbation result due to~\citep{Lei_2015_jour_AnnStat} 
to comment on the deviation of the leading eigenvectors of $L$ from those of $\Lc$. 
A modified version of their result is proved  that incorporates the
row normalization of the eigenvector matrix.
Let $X$ be the matrix of the $k$ leading eigenvectors of $L$,
and $\overline{X}$ be its row normalized version.
We have the following result. Note that since $\delta<1$, the condition of 
Lemma~\ref{lem_random_deviation} is subsumed by the condition stated below.
\begin{lemma}
\label{lem_perturbation}
If $\delta>0$ and $d > {\frac{576\ln n}{\delta^2}}$, then
there is an orthonormal matrix $Q\in\R^{k \times k}$ such that
\begin{equation}
\left\Vert \overline{X} - ZQ\right\Vert_F \leq 
\frac{24}{\delta}\sqrt{\frac{2k n_1 \ln n}{d}}
\label{eq_perturbation_bound}
\end{equation}
with probability at least $(1-\frac{4}{n^2})$.
\end{lemma}

We now derive a bound on the error incurred by the $k$-means step in the algorithm.
Formally, $k$-means  minimizes
$\Vert \overline{X} - S\Vert_F$
over all $S\in\mathcal{M}_{n\times k}(k)$, where
$\mathcal{M}_{n\times k}(r)$ is the set of all $n\times k$ matrices with at most $r$
distinct rows. 
In practice, the rows of $S$ correspond to the centers of obtained clusters.
Achieving a global optimum for this problem is NP-hard. However, there are 
algorithms~\citep{Kumar_2004_conf_FOCS,Ostrovsky_2012_jour_JACM}
that can provide a solution $S^*$ from the above class of matrices such that
\begin{align}
\Vert \overline{X} - S^*\Vert_F \leq \gamma
\underset{S}{\textup{min}} \Vert \overline{X} - S\Vert_F 
\label{eq_suboptimal_kmeans}
\end{align}
for some $\gamma>1$. The factor $\gamma$ depends on the algorithm under consideration.
For instance, $\gamma$ grows with $k$ in the case of~\citep{Kumar_2004_conf_FOCS}.
On the other hand,~\citet{Ostrovsky_2012_jour_JACM} showed that a constant 
factor approximation is possible if the data (rows of $\overline{X}$ in our 
case) is \emph{well-separated}.

To be precise, define $\eta_r(\overline{X})$ to be the minimum of the objective function
when $r$ clusters are found, \ie
\begin{equation}
 \eta_r(\overline{X})  = \min_{S\in\mathcal{M}_{n\times k}(r)} \Vert \overline{X} - S\Vert_F \;.
\end{equation}
The rows of $\overline{X}$ is said to be
$\epsilon$-separated if $\eta_k(\overline{X})  \leq \epsilon \eta_{k-1}(\overline{X})$.
Theorem~4.15 in~\citep{Ostrovsky_2012_jour_JACM} claims that if this condition holds for small 
enough $\epsilon$, then the solution $S^*\in\mathcal{M}_{n\times k}(k)$ obtained from their approximate $k$-means algorithm
satisfies~\eqref{eq_suboptimal_kmeans} with probability $(1- O(\sqrt{\epsilon}))$, where $\gamma$
is given as $\gamma = \sqrt{\displaystyle\frac{1-\epsilon^2}{1-37\epsilon^2}}$.

The following result shows that in our case, the rows of $\overline{X}$ are indeed well-separated.
\begin{lemma}
\label{lem_kmeans_condn}
 If the condition in~\eqref{eq_degree_condn} holds, then the rows of $\overline{X}$ are $\epsilon$-separated
 with $\epsilon = (\ln n)^{-1/2}$.
\end{lemma}

As a consequence of Lemma~\ref{lem_kmeans_condn}, it follows that if $n$ is sufficiently 
large, then the result of~\citep{Ostrovsky_2012_jour_JACM} holds. 
Moreover, one can also observe that for large $n$, we have $\gamma=O(1)$.

Finally, one needs to combine the above results in order to prove Theorem~\ref{thm_main_consistency}. For this, define the set $\Vc_{err} \subset \Vc$ as
\begin{align}
\Vc_{err} = \left\{ i\in \Vc : \Vert S^*_{i\bull} - Z_{i\bull}Q \Vert_2 \geq \frac{1}{\sqrt{2}}
\right\}\;.
\label{eq_W_defn}
\end{align}
\cite{Rohe_2011_jour_AnnStat} used a similar definition 
for the number of incorrectly assigned nodes, and discussed the intuition behind this definition.
In the following result, we formally prove that the nodes that are not in $\Vc_{err}$ are correctly
assigned. We also provide an upper bound on the size of $\Vc_{err}$. 
\begin{lemma}
\label{lem_kmeans}
 Let $i,j\notin\Vc_{err}$ and $S^*_{i\bull} = S^*_{j\bull}$, then $\psi_i=\psi_j$.
 As a consequence, $\err \leq |\Vc_{err}|$.
 In addition, 
 \begin{align*}
  |\Vc_{err}| \leq 4(1+\gamma^2) \Vert \overline{X} - ZQ\Vert_F^2
  \;.
 \end{align*}
\end{lemma}
\noindent
Theorem~\ref{thm_main_consistency} follows by combining the above bound
with~\eqref{eq_perturbation_bound}.

\section{Consistency for Special Cases}
\label{sec_specialcases}
We now study the implications of Theorem~\ref{thm_main_consistency}
for partitioning particular models of uniform and non-uniform hypergraphs.
We also discuss the conditions for identifiability in special cases.

\subsection{Balanced partitions in uniform hypergraph}
\label{subsec_balanced_uniform}
Let the $n$ nodes be divided into $k$ groups such
that each group
contains $\frac{n}{k}$ nodes. We now consider a random $r$-uniform hypergraph
on the nodes generated as follows. Let $p,q\in[0,1]$ be constants with $(p+q)\leq1$,
and $\alpha_{r,n}\in(0,1]$ be the sparsity factor dependent on $n$.
For any $r$ nodes from the same group, there is an edge among them with probability
$\alpha_{r,n}(p+q)$. If all the $r$ nodes do not belong to same group,
then there is an edge with probability $\alpha_{r,n}q$.

In terms of the model in Section~\ref{sec_model}, one can see that $M=r$,
and for all $m<r$, $\alpha_{m,n}=0$. The $r^{th}$ order $k$-dimensional 
tensor $B^{(r)}$ is given by
\begin{align*}
 B^{(r)}_{j_1 j_2 \ldots j_r} = \left\{\begin{array}{ll}
 p+q & \text{if } j_1 = j_2 = \ldots = j_r, \\
 q & \text{otherwise}. \end{array}\right.
\end{align*}
One can see that for $r=2$, this model corresponds to the sparse stochastic block model
considered in~\citep{Lei_2015_jour_AnnStat} with balanced community sizes,
and if $\alpha_{2,n} = 1$, one has the standard four parameter stochastic 
block model~\citep{Rohe_2011_jour_AnnStat}.
The following corollary to Theorem~\ref{thm_main_consistency} shows the 
consistency of Algorithm~\ref{alg_SHP}.

\begin{corollary}
\label{cor_equal_size_uniform}
 In the above model, 
 \begin{equation}
 \delta=\frac{p\alpha_{r,n}n}{rkd}\binom{\frac{n}{k}-2}{r-2},
 \label{eq_delta_uniform}
 \end{equation}
 and hence, the partitions are identifiable for all $p>0$.
Moreover, if 
\begin{equation}
\label{eq_alpha_condn_uniform}
 \alpha_{r,n} \geq C\frac{k^{2r-1} n (\ln n)^2}{\binom{n}{r}}
\end{equation}
 for some absolute constant $C>0$,
 then the conditions in Theorem~\ref{thm_main_consistency} are satisfied, 
 and hence, we have
 \begin{align}
 {\err} = O\left(\frac{k^{2r-2} n^2\ln n}{p^2\alpha_{r,n}\binom{n}{r}}\right) = o(n)
 \label{eq_error_uniform}
 \end{align}
 with probability $(1-o(1))$,
\end{corollary}

The lower bound on $\alpha_{r,n}$ mentioned in 
Corollary~\ref{cor_equal_size_uniform} needs some discussion.
One can verify that in the above model, the expected number of edges lie in 
the range $[q\alpha_{r,n}\binom{n}{r},(p+q)\alpha_{r,n}\binom{n}{r}]$, \ie it is about
$\alpha_{r,n}\binom{n}{r}$ up to a constant scaling.
The lower bound on $\alpha_{r,n}$ specifies that the number of edges must be
at least $\Omega\left(k^{2r-1}n(\ln n)^2\right)$.
This also indicates that for a larger $r$, more edges are required
to ensure the error bound of Corollary~\ref{cor_equal_size_uniform}.
Since, $\alpha_{r,n}\leq1$, one can see that the result is applicable for
$k = {O}(n^{0.5-\epsilon})$ for all $\epsilon > \frac{1}{2(2r-1)}$. 
Even consistency results for graph partitioning require similar condition~\citep{Rohe_2011_jour_AnnStat,Choi_2012_jour_Biometrika}.

A closer look at the condition~\eqref{eq_alpha_condn_uniform} shows that 
if $k$ is constant or increases slowly, $k = O(\ln n)$,  then a sufficient condition for weak 
consistency of Algorithm~\ref{alg_SHP} is 
$\alpha_{r,n} \geq C_r\frac{(\ln n)^{2r+1}}{n^{r-1}}$,
where the constant $C_r$ depends only on $r$.
In case of graph partitioning, this level of sparsity is needed when one relies
on matrix Bernstein inequality.
However, recent results~\citep{Lei_2015_jour_AnnStat} reduced the lower bound by
using sharp concentration bounds for the binary adjacency matrix. 
Corollary~\ref{cor_equal_size_uniform} also indicates that if $k$ increases at a higher rate, for example $k = n^a$,
then consistency can be guaranteed only when hypergraph is more dense.

On the other extreme are uniform hypergraphs encountered in computer vision~\citep{Ghoshdastidar_2014_conf_NIPS,Ghoshdastidar_2015_conf_ICML}
that are usually dense, \ie $\alpha_{r,n} = 1$.
In this case, if $k = O(\ln n)$ then 
$\err = O\left(\frac{(\ln n)^{2r-1}}{n^{r-2}}\right)$.
Thus, the error decreases at a faster rate for $r$-uniform hypergraphs with larger $r$.
In fact, for $r\geq3$, above bound indicates that $\err=o(1)$,  
\ie Algorithm~\ref{alg_SHP} guarantees exact recovery of the partitions for large $n$.

Lastly, we discuss the effect of $\delta$ and the parameters $p,q$ in this setting.
Note that the case $q=0$ is not interesting as there are no edges among different groups,
and hence, the partition can be identified by a simple breadth-first search.
On the other hand, $p=0$ generates a random uniform hypergraph with all identical edges.
Hence, the partitions cannot be identified in this case. 
This can also be seen from~\eqref{eq_delta_uniform}, where $\delta=0$. 
In general, $p$ denotes the gap between the probability of edge occurrence
among nodes from same community and the probability with which nodes from different
communities form an edge.
Since $\delta$ is linear in $p$, one can observe from Theorem~\ref{thm_main_consistency}
that $\err$ varies as $\frac{1}{p^2}$ with $p$.
However, note that the model assumes that 
$p$ does not vary  with $n$, and may be treated as a constant in the asymptotic case.

\subsection{Balanced partitions in non-uniform hypergraph}

We now consider the case of non-uniform hypergraph of range $M$,
where $M$ may vary with $n$.
As in Section~\ref{subsec_balanced_uniform}, assume that $n$ nodes are equally 
split into $k$ groups. Also let $p,q\in(0,1)$ such that $(p+q)\leq1$,
and for $m=2,\ldots,M$, let $B^{(m)}$ be the $m^{th}$-order
symmetric $k$-dimensional tensor with
\begin{align*}
 B^{(m)}_{j_1 j_2 \ldots j_m} = \left\{\begin{array}{ll}
 p+q & \text{if } j_1 = j_2 = \ldots = j_m, \\
 q & \text{otherwise}. \end{array}\right.
\end{align*}
Setting $\alpha_{m,n}\in(0,1]$ as the sparsity factors,
we obtain a model, where the edges appear independently,
and for each $m$, an edge on $m$ nodes from the same group
appears with probability $\alpha_{m,n}(p+q)$.
For any set of $m$ nodes from different groups, there is an edge among them 
with probability $\alpha_{m,n}q$.

Since, the non-unifom hypergraph is a superposition of the $m$-uniform hypergraphs
for $m=2,\ldots,M$, one can easily derive a consistency result
in the non-uniform case by appling Corollary~\ref{cor_equal_size_uniform}
for each of the uniform components. However, 
observe that the number of edges of size $m$ is $\Theta\left(\alpha_{m,n}\binom{n}{m}\right)$,
and hence, the requirement
$\alpha_{m,n}\binom{n}{m} \geq C_m {k^{2m-1} n (\ln n)^2}$
for each $m$ implies that the number of $m$-size edges should increase with $m$.
This contradicts the natural intuition in
existing random models~\citep{Darling_2005_jour_AnnApplProbab},
where the hypergraph contains less edges of higher cardinality.
The same phenomenon is also observed in practice (see Section~\ref{subsec_realsparsity}).
The following consistency result takes this fact into account.
%
\begin{corollary}
\label{cor_equal_size_nonuniform}
The partitions in the above model is identifiable for all $p>0$.
 In addition, let $(\theta_m)_{m=2}^\infty$ be a non-negative sequence independent of $n$,
 and assume that for any $n\in\N$ and $m=2,\ldots,M$,
 the sparsity factor 
 \begin{equation*}
 \alpha_{m,n} = \frac{\theta_m n^a (\ln n)^b}{\binom{n}{m}}
 \end{equation*}
 for some $a\geq1$ and $b\geq2$. There exists an absolute constant $C$, such that, if
 \begin{equation}
 \sum\limits_{m=r}^{M} m\theta_m \leq C\left({\frac{n^{a-1}(\ln n)^{b-2}}{k^{2r-1}}}\right)
 \label{eq_nonuniform_balanced_condn}
 \end{equation}
 for $r= \min\{m: \theta_m>0\}$,
 then $\err = o(n)$ with probability $(1-o(1))$.
\end{corollary}

In the above result, $r$ denotes the smallest size of an edge in the hypergraph.
In practice. $(\theta_m)_{m=2}^\infty$ is a decreasing sequence, 
and hence, the number of $m$-size edges also decreases with $m$.
In particular, if $\theta_2>0$, $\sum\limits_{m=2}^\infty m\theta_m<\infty$, and
$k = O\left({n^{(a-1)/3}(\ln n)^{(b-2)/3}}\right)$,
then Algorithm~\ref{alg_SHP} is weakly consistent.
Thus, if the hypergraph is sparse, \ie $a=1$, consistency is guaranteed only for 
 logarithmic growth in $k$,
whereas larger number of partitions can be consistently detected only in dense hypergraphs.
Observe that the problem gets harder if $r>2$.

\subsection{Identifiability of the partitions}

In the previous two sections, we considered problems where the partitions are identifiable from
$\Lc$. This need not hold for arbitrary model parameters.
We now briefly discuss few cases, which show that the partitions are typically identifiable
under reasonable choice of model parameters. 

\vskip1ex
\noindent {\it Example 1.}
Consider a 3-uniform hypergraph on $n$ nodes. For simplicity assume there are $k\geq3$ 
partitions of equal size. We define  $B^{(3)}$ as follows
\begin{align*}
 B^{(3)}_{j_1 j_2 j_3} = \left\{\begin{array}{ll}
 p_1 & \text{if } j_1 = j_2 = j_3, \\
 p_2 & \text{if exactly two of them are identical,} \\
 p_3 & \text{if } j_1, j_2, j_3 \text{ are all different}.
\end{array}\right.
\end{align*}
for some constants $p_1,p_2,p_3\in[0,1]$.
Observe that the above situation is the most general case provided that the partitions are 
statistically identical.
In this setting, it is easy to see that the following statement holds.
\begin{lemma}
\label{lem_3uniform}
Assume that $n$ is a multiple of $k$. Then $\delta>0$ if and only if
\begin{equation}
(p_2-p_3) + \frac1k(p_1-3p_2+2p_3) - \frac2n(p_1-p_2) > 0.
\label{eq_3uniform_condition}
\end{equation}
In particular, $\delta>0$ when $p_1>p_2>p_3$, or at most one inequality is replaced by equality.
\end{lemma}
Note that the setting of Section~\ref{subsec_balanced_uniform} follows when $p_1>p_2=p_3$,
while the case $p_1=p_2=p_3$ corresponds to a random hypergraph with all edges following
the same law. Obviously, the partitions are not identifiable in the latter case. 
More generally, the order of probabilities $p_1>p_2>p_3$ is intuitive as it implies that an edge
has a larger probability of occurrence if it has more nodes from the same community.
One may compare this observation with the case of graphs, where partitioning based on the 
leading eigenvectors of Laplacian works only when edges within each community occur more 
frequently than edges across communities. The opposite scenario, found in colorable
graphs, requires one to consider eigenvectors corresponding to the other end of the 
spectrum~\citep{Alon_1997_jour_SIAMJComput}.
Moreover, if $p_2>p_3$ and $k$ grows with $n$, one can observe that $\delta$ mostly depends on the gap
$(p_2 - p_3)$, and hence, the error $\err$ is proportional to $\frac{1}{(p_2-p_3)^2}$.

\vskip1ex
\noindent {\it Example 2.}
We now modify the above model by allowing edges of size 2 to be present.
In particular, assume $\alpha_{2,n} = 1$ and $B^{(2)} = I$,
which means all pairwise edges within each community are present, and no two nodes from
different communities form a pairwise edge. In addition, let $\alpha_{3,n} \in[0,1]$ be arbitrary. 
Then, one can observe the following.

\begin{lemma}
\label{lem_23nonuniform}
Assume that $n$ is a multiple of $k$. Then $\delta>0$ if and only if
\begin{equation}
\frac12 + \frac{n\alpha_{3,n}}{3}\left(
(p_2-p_3) + \frac{(p_1-3p_2+2p_3)}{k} - \frac{2(p_1-p_2)}{n}\right) > 0.
\label{eq_23nonuniform_condition}
\end{equation}
\end{lemma}

It is easy to see that if $\alpha_{3,n} = 0$, then the hypergraph is a graph with $k$ 
disconnected components, and hence, the partitions are identifiable. However, even when 
$\alpha_{3,n} = o(\frac1n)$, the pairwise edges eventually dominate and the partitions can be
identified for arbitrary values of $p_1,p_2,p_3$. On the other hand, if $\alpha_{3,n}$ grows 
faster than $\frac1n$ (for instance $\alpha_{3,n} =1$), then the situation is eventually similar
to that of Lemma~\ref{lem_3uniform}. The critical case is $\alpha_{3,n} = \Theta(\frac1n)$,
where the expected number of 2-way and 3-way edges are of similar order.
In this case, \eqref{eq_23nonuniform_condition} suggests that the partitions can be
identified $(\delta>0)$ even when $p_2 < p_3$ provided $p_3$ is sufficiently small.

\vskip1ex
\noindent {\it Example 3.}
In the above cases, we restricted ourselves to communities of equal size. The arguments
also hold for $\frac{n_1}{n_k} = O(1)$. However, if $n_k\ll n_1$ or the probability of edges
vary across different communities, then the second term in~\eqref{eq_delta_charac}
can lead to $\delta\leq0$, or equivalently, may affect the identifiability of the partitions.
To study this effect, we consider the following model for $r$-uniform hypergraphs.

Let $\alpha_{r,n}=1$, and there are $k=2$ partitions of size $s$ and $(n-s)$. We
assume $s=o(n)$, and define $B^{(r)}\in\R^{2\times2\times\ldots\times2}$ as  
\begin{align*}
 B^{(r)}_{j_1 j_2 \ldots j_r} = \left\{\begin{array}{ll}
 1 & \text{if } j_1 = j_2 = \ldots = j_r=1, \\
 \frac12 & \text{otherwise}. \end{array}\right.
\end{align*}
For $r=2$, the model is same as that of a $s$ clique planted in a Erd\"os-R\'enyi graph.
This model presents a high disparity in both community sizes and degree distributions.
We make the following comment on the identifiability of the partitions under this model.

\begin{lemma}
\label{lem_plantedclique}
For a given $r\geq2$, there exists a finite constant $s_r$ such that $\delta>0$ for the above model
for all $s\geq s_r$.
\end{lemma}

Thus, when $s$ grows with $n$, the partitions can be eventually identified from $\Lc$.
The proof of the above result shows that both the terms in~\eqref{eq_delta_charac} decay with $n$,
but the ratio of the first term to the second grows as $\Omega(s)$. We believe that a similar 
observation can be made in more general situations, where this growth rate depends on the 
size of the smallest community. 

In view of the above lemma, it is interesting to know whether Algorithm~\ref{alg_SHP}
is able to detect small cliques in uniform hypergraphs. This is indeed true, but due to the 
generality of the approach, as listed in this paper,  the minimal growth rate for $s$ needed to 
accurately find the partitions from $L$ is not optimal.
More precisely, it is worse by a logarithmic factor in the case of graphs. 
However, Lemma~\ref{lem_plantedclique} shows that
one can use spectral techniques similar to~\citep{Alon_1998_conf_SODA} for finding planted cliques in hypergraphs.

\section{Experimental results}
\label{sec_expt}
In this section, we 
empirically demonstrate that the conditions in Corollaries~\ref{cor_equal_size_uniform}
and~\ref{cor_equal_size_nonuniform} are reasonable.
For this, we consider a number of hypergraphs that have been studied in 
practical problems~\citep{Ghoshal_2009_jour_PhyRevE,Alpert_1998_data_ISPD}.
We also study the performance of Algorithm~\ref{alg_SHP}
in some benchmark clustering problems.


\subsection{Sparsity of real-world hypergraphs}
\label{subsec_realsparsity}
The consistency results in this paper are applicable only under certain restrictions on 
the hypergraph to be partitioned. To be precise, 
Corollaries~\ref{cor_equal_size_uniform} and~\ref{cor_equal_size_nonuniform} hold
when the sparsity of the hypergraph is above a certain threshold. We study the practicability
of the conditions in the case of real-world hypergraphs.
We consider two types of applications -- \emph{folksonomy}, where the underlying model 
is a 3-uniform hypergraph, and \emph{circuit design}, which involves non-uniform
hypergraph partitioning.

\begin{figure}[ht]
\begin{tabular}{cc}
\rotatebox{90}{~~~~~$|\Ec|$ and $n(\ln n)^2$  in log scale} &
\includegraphics[width=0.5\textwidth,clip=true,trim=19mm 12mm 17mm 9mm]{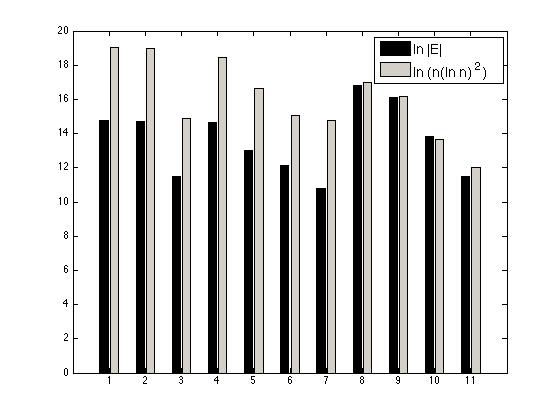}
\\& Networks \\
\end{tabular}
\caption{Bar plot for $|\Ec|$ and $n(\ln n)^2$ in logarithmic scale for 11 folksonomy networks.}
\label{fig_folksonomy}
\end{figure}

To study the nature of hypergraphs in folksonomy, we consider 11 networks from
KONECT, HetRec'2011 and MovieLens\footnote{
The HetRec'2011 and MovieLens datasets are maintained by the GroupLens research group,
and are available at: \texttt{http://grouplens.org/} \newline
KONECT refers to the Koblenz network collection: \texttt{http://konect.uni-koblenz.de/} 
}. 
Each network is a tri-partite 3-uniform hypergraph containing three types of nodes --
user, resource and annotation. Each edge is an entry in the database that occurs when
an user describes a certain resource by a particular tag or rating. 
The number of nodes vary between 2630 to $9.8\times10^5$.
Assuming that $k = O(1)$,
the sufficient condition in Corollary~\ref{cor_equal_size_uniform} requires that the number of 
edges in a 3-uniform hypergraph grows as $\Omega\left(n (\ln n)^2\right)$.
 In Figure~\ref{fig_folksonomy}, we compare the number of edges $|\Ec|$ 
 with $n(\ln n)^2$ for above networks.
 We observe that in few cases (last four in Figure~\ref{fig_folksonomy}), these quantities are similar,
 whereas for the remaining networks, $|\Ec|$ is smaller by a nearly constant factor. 

\begin{figure}[ht]
\begin{tabular}{cc}
\rotatebox{90}{~~~~~~~~~~~~~Estimate of $\theta_m$} &
\includegraphics[width=0.5\textwidth,clip=true,trim=18mm 11mm 16mm 8mm]{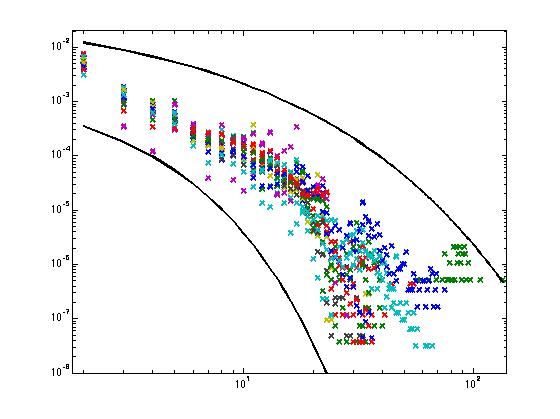}
\\& Size of edge, $m$ \\
\end{tabular}
\caption{Scatter plot for estimated $\theta_m=\frac{|\Ec_m|}{n(\ln n)^{2}}$
versus $m$ for the 18 circuits.
Plot for each circuit is shown in a different color.
The bounding curves correspond to the functions $0.05\exp(-m^{0.5})$ from above
and  $0.002\exp(-m^{0.8})$ from below.}
\label{fig_circuit}
\end{figure}

The next study is related to non-uniform hypergraphs that are encountered in 
circuit partitioning. We consider 18 circuits from the 
ISPD98 circuit benchmark suite~\citep{Alpert_1998_data_ISPD}. 
From a hypergraph view, the components of the circuit
are the nodes of the hypergraph, while the multi-way connections among them
are the edges. These networks are also sparse as
the number of nodes vary from 
$1.27\times10^4$ to $2.1\times10^5$, while the number of edges range between
$1.4\times10^4$ to $2\times10^5$. 
Moreover, these networks contain relatively large number of edges of sizes
2 or 3, and the number of edges of size $m$ gradually decreases with $m$.
%
We assume $a=1$, $b=2$, and ignoring constant factors,
we estimate $\theta_m$ as
$\theta_m = \frac{|\Ec_m|}{n(\ln n)^2}$, where $\Ec_m$
is the set of edges of size $m$ in the network.
Figure~\ref{fig_circuit} shows a plot of this quantity as a function of $m$ for 
different networks. 
%
We find that the estimate of $\theta_m$ is bounded  
by exponentially decaying functions, and hence, one can argue that 
$\sum_m m\theta_m<\infty$.

\subsection{Experiments on benchmark problems}

Partitioning the networks discussed in the previous section is an interesting problem.
However, for such networks, the underlying partitions are not known, and hence,
for these networks, the performance of Algorithm~\ref{alg_SHP} cannot be measured 
in terms of the number of incorrectly assigned nodes.
So, we consider benchmark problems for categorical data clustering,
where the true partitions are known.
%
Here, one needs to group instances of a database, each described by
a number of categorical attributes. Two such benchmark databases include the 
\emph{1984 US Congressional Voting Records} and the \emph{Mushroom Database} 
available at the UCI repository~\citep{Lichman_2013_dataset_UCI}.
The first set contains votes of 435 Congress men on 16 issues. 
The task is to group the Congress men into Democrats and Republicans
based on whether they voted for or against the issue, or abstained their votes.
The mushroom database contains information about 22 features
of 8124 varieties of mushrooms. Based on the categorical features,
one needs to separate the edible varieties from the poisonous ones.
Thus, both databases have two well-defined partitions.

%
One may consider the instances 
of the database as the nodes of the hypergraph.
For each possible value of each attribute, an edge is 
considered among all instances that take the particular
value of the attribute. This generates a sparse non-uniform hypergraph
that can be partitioned to obtain the clusters.
%
Table~\ref{tab_categorical}
compares the performance of Algorithm~\ref{alg_SHP}
with some popular categorical clustering algorithms.
We also study the performance when the hypergraph partitioning
is done using a multi-level approach (hMETIS)~\citep{Karypis_2000_jour_VLSI} or
eigen-decomposition of the normalized Laplacian obtained from 
clique expansion~\citep{Rodriguez_2002_jour_LinMultilinAlg}. 
The error is measured as $\frac{1}{n}\err$.
Table~\ref{tab_categorical} shows that Algorithm~\ref{alg_SHP} performs 
quite well compared to other methods.
%
\begin{table}[ht]
\centering
 \caption{Fraction of nodes incorrectly assigned by different clustering algorithms.
 The results for ROCK, COOLCAT and LIMBO are taken from~\citep{Andritos_2004_conf_CEDT}.}
 \label{tab_categorical}
 \begin{tabular}{ccccccc}
  \hline
  Database & ROCK & COOLCAT & LIMBO & hMETIS 
  & Clique & Algorithm~\ref{alg_SHP} \\
  \hline
  Voting & 0.16 & 0.15 & 0.13 & 0.24 & 0.12 & 0.12\\
  Mushroom & 0.43 & 0.27 & 0.11 & 0.48 & 0.11 & 0.11 \\
  \hline
 \end{tabular}
\end{table}

\section{Conclusion}
\label{sec_conclusion}
The primary focus of this work was to study the consistency of hypergraph partitioning
in the presence of a planted structure in the hypergraph. 
This is achieved by considering a model for random hypergraphs that extends the 
stochastic block model in a natural way. 
The algorithm studied in this work is quite simple,
where one essentially reduces a given hypergraph to a graph with weighted adjacency matrix
given by~\eqref{eq_A_defn}, and then performs spectral clustering on this graph.
Our analysis mainly relies on a matrix concentration inequality that was previously used
to derive concentration bounds for the Laplacian matrix of sparse random
graphs~\citep{Chung_2011_jour_EJComb}.
We also establish that the $k$-means step indeed achieves a constant factor approximation
with probability $(1-o(1))$. This question had remained unanswered for a long time in the 
spectral clustering literature. 

\subsection*{Note on the optimality of our result}
Theorem~\ref{thm_main_consistency} is quite similar, in spirit, to the existing
results in the block model literature, for instance~\citep{Lei_2015_jour_AnnStat},
where it is shown that spectral clustering is weakly consistent when the expected degree
of any node is $\Omega(\ln n)$, or equivalently the edge density
$\alpha_{2,n} = \Omega(\frac{\ln n}{n})$. One can easily see from
Corollary~\ref{cor_equal_size_uniform} that our result is not optimal, at least in the case of 
graphs. The primary factor contributing to this difference is a sharp concentration 
result~\citep{Friedman_1989_conf_STOC} that holds for the sparse binary adjacency matrix.
While such a sharp bound may not hold for a weighted adjacency 
matrix, we do wonder whether one can consider the following approach.
\vskip0.5ex
\noindent {\it Question:} 
Viewing a hypergraph as a collection of $m$-uniform hypergraphs,
one can represent the adjacencies as a collection of $m$-way binary tensors for varying $m$.
Does a generalization of~\citep{Friedman_1989_conf_STOC} hold for sparse binary tensors?
If so, then what is its implication on the allowable sparsity for community detection in hypergraphs?
\vskip0.5ex \noindent
One can show that for dense tensors, the operator norm (equivalently, largest eigenvalue for matrices)
does concentrate similar to the matrix case~\citep{Ghoshdastidar_2015_conf_ICML},
but the sparse case has not been studied yet.

Considering the most sparse regime for community detection~\citep{Decelle_2011_jour_PhyRevE}, 
it is now known that spectral techniques based on eigenvectors of suitably
defined matrices work even for graphs with density $\alpha_{2,n}= \Theta(\frac1n)$ 
\citep{Le_2015_arxiv,Krzakala_2013_jour_PNAS}.
To this end, the following problem is quite interesting.
\vskip0.5ex
\noindent {\it Question:} 
What is an appropriate extension of the regularized adjacency matrix~\citep{Le_2015_arxiv}
and the non-backtracking matrix~\citep{Krzakala_2013_jour_PNAS} in the case of hypergraphs? 
More generally, what is the algorithmic barrier for community detection in hypergraphs?
\vskip0.5ex \noindent 
Phase transitions in uniform hypergraphs have been studied in the 
literature~\citep{Achlioptas_2008_conf_FOCS,Panangiotou_2012_conf_STOC}, 
and thresholds for 2-colorability and boolean satisfiability are known up to constant factors.
However, the case of non-uniform hypergraphs still remains unexplored.

\subsection*{Extensions of our results}
One can observe that both the model and the analysis can be further extended to more
general situations.
For instance, one often encounters weighted hypergraphs in practical applications~\citep{Ghoshdastidar_2014_conf_NIPS}, where every edge $e$ has a weight $w(e)$ associated with it.
In our random model, we assumed $w(e)$ to be a Bernoulli random variable.
A direct extension to the weighted hypergraphs is obtained by allowing $w(e)$ to take real
values. To this end, we note that our results are only based on the first two moments of $w(e)$.
Hence, if we restrict $w(e)\in[0,1]$ and assume that first moment is same as that 
of the Bernoulli variables in our model, then Theorem~\ref{thm_main_consistency}
holds even in this setting.

In the case of planted graphs, the stochastic block model has been extended to 
account for factors such as degree heterogenity or overlapping communities~\citep{Lei_2015_jour_AnnStat,Zhang_2014_arxiv}. 
Similar modifications of the hypergraph model is an interesting extension.  
However, it also seems possible that some information, such as 
community overlap, may be lost when the edge information is `compressed' into
the hypergraph Laplacian. Hence, one may have to consider
spectral properties of the incidence matrix $H$ or other
alternatives for the Laplacian in~\eqref{eq_hyperlap}.

While we restricted our discussions to a popular hypergraph partitioning approach,
the results can be extended to variants of Algorithm~\ref{alg_SHP}.
For instance, one may use the eigenvectors of the weighted adjacency matrix $A$
instead of the Laplacian $L$. Minor modifications to Theorem~\ref{thm_main_consistency}
can guarantee the consistency of such an approach.
Moreover, Theorem~\ref{thm_main_consistency} is based on a theoretical result
of approximate $k$-means~\citep{Ostrovsky_2012_jour_JACM}. 
As mentioned in Section~\ref{sec_main_result},  we need to assume $n$ to be sufficiently 
large in order to ensure that the $k$-means step provides a near optimal solution.
Alternatively, one could also use the greedy clustering algorithm of \citet{Gao_2015_arxiv}
that may alleviate the condition on $n$.

It also is known that one can iteratively refine the solution of a spectral algorithm to exactly 
recover the partitions~\citep{Vu_2014_arxiv,Lei_2014_arxiv}. Such an approach usually
constructs an embedding of the nodes based on the adjacency matrix.
We believe that similar results will hold for hypergraphs if one constructs the embedding using
the weighted adjacency matrix $A$ defined in~\eqref{eq_A_defn}.

We noted in Lemma~\ref{lem_plantedclique} that Algorithm~\ref{alg_SHP} can be used
to find a planted clique in a random hypergraph. For a more optimal result, 
one could possibly extend the approach of~\citep{Alon_1998_conf_SODA} to the case of
hypergraphs. To this end, it is interesting to note that the hypergraph clique problem 
is often encountered in computer vision applications~\citep{Ghoshdastidar_2014_conf_NIPS}.
Thus, several variations of the hypergraph partitioning problem surface in engineering 
applications. 

This work explores into the theoretical analysis of hypergraph partitioning, and
provides the first step for expanding the extensive studies on 
planted graphs to the case of hypergraphs.


\appendix
\section{Proofs for Corollaries in Section~\ref{sec_main_result}}
The proofs are similar to that of Theorem~\ref{thm_main_consistency}.
To prove Corollary~\ref{cor_prob1_kmeans}, one needs to use the lemmas in
Section~\ref{sec_main_result}, except Lemma~\ref{lem_kmeans_condn}.
On the other hand, Corollary~\ref{cor_constprob_kmeans} follows when we use
Lemma~\ref{lem_kmeans_condn} with constant $\epsilon>0$. The condition
$\epsilon<0.015$ follows immediately from the requirement of Lemma~4.11 in~\citep{Ostrovsky_2012_jour_JACM}. 

\section{Proofs for Lemmas in Section~\ref{sec_main_result}}

\subsection{Proof of Lemma~\ref{lem_expected1}}
We observe that for $i\neq j$,
\begin{align}
 \Ac_{ij} 
 = \sum_{\ell=1}^{\beta_{M}} \frac{\E[h_\ell]}{|\xi_\ell|} (a_{\xi_\ell})_i (a_{\xi_\ell})_j
 &= \sum_{m=2}^{M} \;\sum_{\substack{\ell:|\xi_\ell|=m,\\i,j\in\xi_\ell}} \frac{\E[h_\ell]}{m}
 \nonumber
 \\&= \sum_{m=2}^{M} \; \sum_{\substack{i_3<i_4<\ldots<i_m,\\i,j\notin\{i_3,\ldots,i_m\}}} 
 \frac{1}{m}\alpha_{m,n} B^{(m)}_{\psi_i\psi_j\psi_{i_3}\ldots\psi_{i_m}} \;.
 \label{eq_Ac_expn}
\end{align}
The last equality follows by noting that for every $\xi_\ell$ such that $|\xi_\ell|=m$
and $\xi_\ell\ni i,j$, we can write $\xi_\ell$ as $\xi_\ell = \{i,j,i_3,\ldots,i_m\}$,
where the nodes $i,j,i_3,...,i_m$ are distinct.
It is interesting to note that the above sum remains same if $i,j$
are replaced by some $i',j'$ such that $\psi_i = \psi_{i'}$ 
and $\psi_j = \psi_{j'}$. This is true since the terms in~\eqref{eq_Ac_expn} depend on
$\psi_i,\psi_j$ instead of $i,j$.
%
This observation motivates us to define the matrix $G\in\R^{k\times k}$
such that for any $i,j\in\Vc$, $i\neq j$,
\begin{align}
 G_{\psi_i \psi_j} = \sum_{m=2}^{M} \sum_{\substack{i_3<i_4<\ldots<i_m,\\i',j'\notin\{i_3,\ldots,i_m\}}} 
 \frac{1}{m}\alpha_{m,n} B^{(m)}_{\psi_{i'}\psi_{j'}\psi_{i_3}\ldots\psi_{i_m}} \;,
\end{align}
where $i',j'$ are arbitrary nodes satisfying $\psi_i = \psi_{i'}$ 
and $\psi_j = \psi_{j'}$.
Hence, one can write $\Ac_{ij} = (ZGZ^T)_{ij}$ for all $i\neq j$, 
where $Z$ is the assignment matrix.
However, 
\begin{align*}
 \Ac_{ii} = \sum_{\ell:i\in\xi_\ell} \frac{\E[h_\ell]}{|\xi_\ell|} \neq G_{\psi_i\psi_i}.
\end{align*}
So, one can write the matrix $\Ac$ as
$\Ac = ZGZ^T - J$,
where $J\in\R^{n\times n}$ is a diagonal matrix defined as $J_{ii} = G_{\psi_i \psi_i} - \Ac_{ii}$.
We also note that for $i,i'$ in the same group, \ie $\psi_i = \psi_{i'}$,
we have
$\Dc_{ii} = \Dc_{i'i'}$ and $J_{ii} = J_{i'i'}$.
So we can define matrices $\Dt,\Jt\in\R^{k\times k}$ diagonal
such that $\Dc_{ii} = \Dt_{\psi_i \psi_i}$ and $J_{ii} = \Jt_{\psi_i \psi_i}$
for all $i\in\Vc$.
It is easy to see that $\Dc Z = Z\Dt$ and $JZ = Z\Jt$.

Using above definitions, we now characterize the 
eigenpairs of the matrix $\Dc^{-1/2}\Ac\Dc^{-1/2}$. 
To this end, note that
$(\lambda,v)$ is an eigenpair of $\Lc$ if and only if
 $((1-\lambda),v)$ is an eigenpair of $\Dc^{-1/2}\Ac\Dc^{-1/2}$.
Hence, it suffices to consider
the eigenvalues of $\Dc^{-1/2}\Ac\Dc^{-1/2}$, and 
their corresponding eigen spaces.

First, observe that since $G\in\R^{k\times k}$,
$\Ac$ is composed of a matrix of rank at most $k$ that is perturbed by 
the diagonal matrix $J$. We show that the orthonormal basis for the range space of $ZGZ^T$
are the eigenvectors that are of interest to us. For this,
consider $\mathcal{G} = (\Dt^{-1}Z^TZ)^{1/2}G(Z^TZ\Dt^{-1})^{1/2} - \Jt\Dt^{-1} \in\R^{k \times k}$,
and suppose its eigen-decomposition is given by $\mathcal{G} = U\Lambda_1U^T$,
where $U\in\R^{k \times k}$ contains the orthonormal eigenvectors
and $\Lambda_1\in\R^{k \times k}$ is a diagonal matrix of eigenvalues of $\mathcal{G}$.
Defining $\Xc = Z(Z^T Z)^{-1/2} U \in\R^{n\times k}$, 
we can write  that
\begin{align*}
 \Dc^{-1/2}\Ac\Dc^{-1/2} \Xc
 &= \Dc^{-1/2}(ZGZ^T - J)\Dc^{-1/2} Z(Z^T Z)^{-1/2} U 
 \\&= \Dc^{-1/2} (ZG(Z^T Z)^{1/2} - Z(Z^T Z)^{-1/2} \Jt) \Dt^{-1/2} U
 \\&= Z(Z^T Z)^{-1/2} \mathcal{G} U
 \\&= Z(Z^T Z)^{-1/2} U \Lambda_1 = \Xc\Lambda_1,
\end{align*}
which implies that the columns of $\Xc$ are the eigenvectors
of $\Dc^{-1/2}\Ac\Dc^{-1/2}$ corresponding to the $k$ eigenvalues in $\Lambda_1$.
Alternatively, the columns of $\Xc$ are the eigenvectors
of $\Lc$ corresponding to the $k$ eigenvalues in $(I-\Lambda_1)$.
Note that the above equalities are derived by repeated use of the facts that diagonal 
matrices commute and $\Dc Z = Z\Dt$, $JZ = Z\Jt$.
Also, since $U$ is orthonormal, it is easy to verify that the columns 
of $\Xc$ are orthonormal.

\subsection{Proof of Lemma~\ref{lem_expected}}
We continue from the proof of Lemma~\ref{lem_expected1}. 
Note that we need to derive conditions
under which $\Xc$ contain the leading eigenvectors of $\Lc$. Equivalently,
we need to show that the eigenvalues in $\Lambda_1$ are strictly larger than other eigenvalues 
of $\Dc^{-1/2}\Ac\Dc^{-1/2}$.

Since, $\Dc^{-1/2}\Ac\Dc^{-1/2}$ is symmetric and hence, diagonalizable,
we can conclude that remaining eigenvectors of the matrix are orthogonal 
to columns of $\Xc$. 
Let the columns of $Y\in\R^{n \times (n-k)}$ be the matrix of the remaining 
orthonormal eigenvectors of $\Dc^{-1/2}\Ac\Dc^{-1/2}$,
with corresponding eigenvalues given by the diagonal matrix $\Lambda_2\in\R^{(n-k)\times(n-k)}$. 
So $Y^TZ(Z^T Z)^{-1/2} U =0$. 
Due to the non-singularity of $Z^TZ$ or $U$, it follows that $Z^TY=0$, and
\begin{align*}
 Y\Lambda_2 &= \Dc^{-1/2}\Ac\Dc^{-1/2}Y
 = -\Dc^{-1}JY ,
\end{align*}
that is, the columns of $Y$ are eigenvectors of $(-\Dc^{-1}J)$.
Further, since $\Dc^{-1}J$ is diagonal,
the eigenvalues in $\Lambda_2$ are a subset of the entries of $(-\Dc^{-1}J)$.
Thus, to ensure that $\Xc$ are the leading eigenvectors,
one needs to ensure $\min_i (\Lambda_1)_{ii} > \max_i (\Lambda_2)_{ii}$,
and hence, one may define $\widetilde\delta$ as the eigen-gap,
\begin{equation}
 \widetilde\delta = \min_{1\leq i\leq k} (\Lambda_1)_{ii} - \max_{1\leq i\leq (n-k)} (\Lambda_2)_{ii}.
 \label{eq_eigengap_defn}
\end{equation}

Hence, the condition $\widetilde\delta>0$ ensures that columns of $\Xc$ are leading
eigenvectors of $\Lc$.
Though the above definition of $\widetilde\delta$ suffices,
it cannot be easily verified for a given model. Below, we
show that $\widetilde\delta \geq \delta$, where the latter is as defined in~\eqref{eq_delta_charac}.
Note that 
\begin{align*}
 \max_{1\leq i\leq (n-k)} (\Lambda_2)_{ii} \leq \max_{1\leq i\leq n} \left(-\frac{J_{ii}}{\Dc_{ii}}\right) 
 = \min_{1\leq i\leq n} \frac{J_{ii}}{\Dc_{ii}} 
\end{align*}
On the other hand, using Weyl's inequality, we have 
\begin{align*}
 \min_{1\leq i\leq k} (\Lambda_1)_{ii} 
 &= \lambda_{\min}(\mathcal{G})
 \geq \lambda_{\min}((\Dt^{-1}Z^TZ)^{1/2}G(Z^TZ\Dt^{-1})^{1/2}) - \Vert\Jt\Dt^{-1}\Vert_2\;,
\end{align*}
where $\lambda_{\min}(\mathcal{G})$ denotes the minimum eigenvalue of $\mathcal{G}$.
The inequality follows by viewing $\mathcal{G}$ as the matrix
$(\Dt^{-1}Z^TZ)^{1/2}G(Z^TZ\Dt^{-1})^{1/2}$ perturbed by $-\Dt^{-1}\Jt$.
To simplify further, we note 
\begin{align*}
 \Vert\Jt\Dt^{-1}\Vert_2 = \max_{1\leq i\leq k} \frac{\Jt_{ii}}{\Dt_{ii}} = \max_{1\leq i\leq n} \frac{J_{ii}}{\Dc_{ii}} ,
\end{align*}
and using Rayleigh's principle, one can show that
\begin{align*}
 \lambda_{\min}((\Dt^{-1}Z^TZ)^{1/2}G(Z^TZ\Dt^{-1})^{1/2}) \geq \lambda_{\min}(G)
 \min_{1\leq i\leq k} \frac{(Z^TZ)_{ii}}{\Dt_{ii}}.
\end{align*}
Combining the above bounds, we conclude that $\widetilde\delta\geq\delta$.
Here, we use the observation that $(Z^TZ)_{jj}$ equals the size of the $j^{th}$ community.
Thus, $\delta>0$ is a sufficient condition for the claim of the lemma.

%

\subsection{Proof of Lemma~\ref{lem_random_deviation}}
Define $\Lh = I - \Dc^{-1/2}A\Dc^{-1/2}$.
Note that 
\begin{align}
 \Vert L - \Lc\Vert_2 \leq \Vert L - \Lh\Vert_2 + \Vert\Lc - \Lh\Vert_2.
 \label{eq_L_Lc_bound_proof}
\end{align}
We deal with the two terms separately. First, we show that if $d>9\ln n$, then
\begin{align}
 \P\left( \Vert \Lc - \Lh\Vert_2 \geq  3\sqrt{\frac{\ln n}{d}} \right) \leq \frac{2}{n^2} \;.
 \label{eq_LcLh_bound}
\end{align}
To prove~\eqref{eq_LcLh_bound}, we note that
\begin{align*}
 \Lc - \Lh &= \Dc^{-1/2}(A - \Ac)\Dc^{-1/2}
 \\&= \sum_\ell (h_\ell - \E[h_\ell]) \frac{1}{|\xi_\ell|} \Dc^{-1/2}a_{\xi_\ell} a_{\xi_\ell}^T \Dc^{-1/2}\;.
\end{align*}
Denoting, each matrix in the sum as $Y_\ell$, it is easy to see that $\{Y_\ell\}_\ell$
are independent with $\E[Y_\ell]=0$. 
Hence, we can apply matrix Bernstein inequality~\citep{Chung_2011_jour_EJComb} to obtain
\begin{align}
 \P&\left( \Vert \Lc - \Lh\Vert_2 \geq  3\sqrt{\frac{\ln n}{d}} \right) 
 = \P\left( \left\Vert \textstyle\sum\limits_\ell Y_\ell \right\Vert_2 \geq  3\sqrt{\displaystyle\frac{\ln n}{d}} \right) 
 \nonumber
 \\&\hspace{15mm}\leq 
 2n\exp\left( \frac{-\displaystyle\frac{9\ln n}{d}}{
 2\left\Vert \sum\limits_\ell \var(Y_\ell) \right\Vert_2 + \displaystyle\frac{2}{3}\sqrt{\frac{9\ln n}{d}}\max_\ell \Vert Y_\ell\Vert_2
 }\right)\;,
 \label{eq_LcLh_bound1} 
\end{align}
where 
\begin{align*}
 \var(Y_\ell) &= \E[Y_\ell^2]
 = \var(h_\ell) \frac{(a_{\xi_\ell}^T \Dc^{-1} a_{\xi_\ell})}{|\xi_\ell|^2} \Dc^{-1/2} a_{\xi_\ell}a_{\xi_\ell}^T \Dc^{-1/2}\;.
\end{align*}
Thus,
\begin{align*}
 \sum_\ell \var(Y_\ell) = \Dc^{-1/2} \left(\sum_\ell 
 \var(h_\ell) \frac{(a_{\xi_\ell}^T \Dc^{-1} a_{\xi_\ell})}{|\xi_\ell|^2}  a_{\xi_\ell}a_{\xi_\ell}^T \right) \Dc^{-1/2} \;.
\end{align*}
Note that for any matrix $B$, $\Dc^{-1/2}B\Dc^{-1/2}$ and $\Dc^{-1}B$ have same eigenvalues,
and hence, using Gerschgorin's theorem~\citep{Stewart_1990_book_AP}, one has
\begin{align*}
 \left\Vert\sum_\ell \var(Y_\ell) \right\Vert_2 
 &\leq \max_{1\leq i \leq n} \frac{1}{\Dc_{ii}} \sum_{j=1}^n \left(
 \sum_\ell\var(h_\ell) \frac{(a_{\xi_\ell}^T \Dc^{-1} a_{\xi_\ell})}{|\xi_\ell|^2}  a_{\xi_\ell}a_{\xi_\ell}^T \right)_{ij}
 \\&= \max_{1\leq i \leq n} \frac{1}{\Dc_{ii}}
 \sum_\ell\var(h_\ell) \frac{(a_{\xi_\ell}^T \Dc^{-1} a_{\xi_\ell})}{|\xi_\ell|^2}  (a_{\xi_\ell})_i \sum_{j=1}^n (a_{\xi_\ell})_j \;.
\end{align*}
Observing that $a_{\xi_\ell}^T \Dc^{-1} a_{\xi_\ell} \leq \frac{a_{\xi_\ell}^T a_{\xi_\ell}}{d}$ 
and $|\xi_\ell| = \sum_j (a_{\xi_\ell})_j = a_{\xi_\ell}^T a_{\xi_\ell}$, we have
\begin{align*}
 \left\Vert\sum_\ell \var(Y_\ell) \right\Vert_2 
 &\leq \frac{1}{d} \max_{1\leq i \leq n} \frac{1}{\Dc_{ii}} 
 \sum_\ell\var(h_\ell) (a_{\xi_\ell})_i
 \leq \frac{1}{d}
\end{align*}
since $\var(h_\ell) = \E[h_\ell](1 - \E[h_\ell]) \leq \E[h_\ell]$.
Similarly, one can also compute
\begin{align*}
 \Vert Y_\ell \Vert_2 
 &\leq |h_\ell - \E[h_\ell]| \frac{1}{|\xi_\ell|} \Vert\Dc^{-1/2}a_{\xi_\ell} a_{\xi_\ell}^T \Dc^{-1/2}\Vert_2
 \leq \frac{|a_{\xi_\ell}^T \Dc^{-1} a_{\xi_\ell}|}{|\xi_\ell|}
 \leq \frac{1}{d},
\end{align*}
where second inequality holds since
$h_\ell\in\{0,1\}$ and $\Dc^{-1/2}a_{\xi_\ell} a_{\xi_\ell}^T \Dc^{-1/2}$ is a rank-1 matrix.
Substituting above bounds in~\eqref{eq_LcLh_bound1} and noting that $\frac{9\ln n}{d}<1$, we have
\begin{align*}
 \P\left( \Vert \Lc - \Lh\Vert_2 \geq  3\sqrt{\frac{\ln n}{d}} \right) 
 &\leq 
 2n\exp\left( -\frac{9\frac{\ln n}{d}}{ \frac{2}{d} + \frac{1}{d} }\right)
 = \frac{2}{n^2}\;,
\end{align*}
which proves~\eqref{eq_LcLh_bound}. 
To bound the other term in~\eqref{eq_L_Lc_bound_proof}, we note that
\begin{align*}
 \Vert L - \Lh\Vert_2
 &\leq \Vert \Dc^{-1/2}A\Dc^{-1/2} - D^{-1/2}AD^{-1/2}\Vert_2
 \\&\leq \Vert (\Dc^{-1/2} - D^{-1/2})A\Dc^{-1/2} + D^{-1/2}A(\Dc^{-1/2} - D^{-1/2})\Vert_2
 \\&\leq \Vert (\Dc^{-1}D)^{1/2} - I\Vert_2\Vert (D\Dc^{-1})^{1/2}\Vert_2 + \Vert(\Dc^{-1}D)^{1/2} - I\Vert_2 .
\end{align*}
In above, we use the fact that $D_{ii} = \sum_j A_{ij}$ to conclude that $\Vert D^{-1/2}AD^{-1/2}\Vert_2 = 1$.
Note that $\Dc^{-1}D$ is a diagonal matrix with non-negative diagonal entries, and hence,
\begin{align*}
 \Vert (\Dc^{-1}D)^{1/2} - I\Vert_2 
 = \max_{1\leq i \leq n} \left| \sqrt{\frac{D_{ii}}{\Dc_{ii}}} - 1\right|
 \leq \max_{1\leq i \leq n} \left| {\frac{D_{ii}}{\Dc_{ii}}} - 1\right|,
\end{align*}
where the inequality follows from the fact that $|\sqrt{x}-1| \leq |x-1|$ for all $x\geq0$.
We now claim that for all $i=1,\ldots,n$,
\begin{align}
 \P\left( |D_{ii}-\Dc_{ii}| > 3\Dc_{ii}\sqrt{\frac{\ln n}{d}}\right) \leq \frac{2}{n^3}\;.
 \label{eq_degree_bound}
\end{align}
Hence, with probability at least $(1 - \frac{2}{n^2})$,
\begin{align*}
 \max_{1\leq i \leq n} \left| {\frac{D_{ii}}{\Dc_{ii}}} - 1\right| \leq 3\sqrt{\frac{\ln n}{d}} \;.
\end{align*}
From above and the relation
$\Vert (D\Dc^{-1})^{1/2}\Vert_2 \leq 1 + \Vert (D\Dc^{-1})^{1/2}-I\Vert_2$, we have 
\begin{align*}
 \Vert L - \Lh\Vert_2 \leq \frac{9\ln n}{d} + 6\sqrt{\frac{\ln n}{d}} 
 \leq 9\sqrt{\frac{\ln n}{d}} \;
\end{align*}
where the last inequality holds since $3\sqrt{\frac{\ln n}{d}} < 1$.
The lemma follows by combining above bound with~\eqref{eq_LcLh_bound}.

Finally, we prove~\eqref{eq_degree_bound}. Since,
$D_{ii} = \displaystyle\sum_\ell h_\ell (a_{\xi_\ell})_i = \displaystyle\sum_{\ell:i\in\xi_\ell} h_\ell$,
we use Bernstein inequality to write
\begin{align*}
 \P\left( |D_{ii}-\Dc_{ii}| > 3\Dc_{ii}\sqrt{\frac{\ln n}{d}}\right)
 &= \P\left( \left|\sum_{\ell:i\in\xi_\ell} (h_\ell - \E[h_\ell])\right| > 3\Dc_{ii}\sqrt{\frac{\ln n}{d}}\right)
 \\&\leq 2\exp\left(\frac{-\displaystyle\frac{9\Dc_{ii}^2\ln n}{d}}{
 2\displaystyle\sum_{\ell:i\in\xi_\ell} \var(h_\ell) + 2\Dc_{ii}\sqrt{\frac{\ln n}{d}}}\right)
 \\&\leq 2\exp\left(-\frac{3\Dc_{ii}\ln n}{d}\right)
\end{align*}
for $d>9\ln n$. Since, $\Dc_{ii}\geq d$, we obtain~\eqref{eq_degree_bound}.

\subsection{Proof of Lemma~\ref{lem_perturbation}}
To perform a valid row normalization of $X$, first we need to ensure
that the rows of $X$ are non-zero with high probability.
From~\eqref{eq_degree_bound}, we see that for all $i\in\Vc$,
\begin{align*}
 \P\left(D_{ii} \geq \Dc_{ii}\left(1-3\sqrt{\frac{\ln n}{d}}\right)\right)
 \geq 1 - \frac{2}{n^3}\;,
\end{align*}
where the second term in lower bound of $D_{ii}$ is smaller than 1 for $d>9\ln n$.
In addition, we know $\Dc_{ii} \geq d > 9\ln n$. Combining above with union bound,
we can say that $\P(\min_i D_{ii}>0) \geq 1 -\frac{2}{n^2}$,
\ie the hypergraph contains no disconnected node with probability at least $(1-\frac{2}{n^2})$.

One can verify the following properties for $L$ in~\eqref{eq_hyperlap},
which can be derived using arguments similar to those for graph Laplacian~\citep{vonLuxburg_2007_jour_StatComp}.
\begin{itemize}
 \item $L$ is positive semi-definite with eigenvalues in $[0,2]$.
 \item The multiplicity of 0 eigenvalue of $L$ is equal to the number of connected components of the hypergraph.
 \item Provided that $D_{ii}>0$ for all $i$, for each zero eigenvalue of $L$, there is an eigenvector
       with non-zero coordinates for one connected component.
\end{itemize}

From the condition $\delta>0$ in Lemma~\ref{lem_expected}, there is a strictly positive eigen-gap between
the $k^{th}$ and $(k+1)^{th}$ eigenvalues of $\Lc$, and hence, $\Lc$ has atmost
$k$ zero eigenvalues. Denoting $\lambda_i(L),\lambda_i(\Lc)$ as the $i^{th}$ smallest 
eigenvalues of $L,\Lc$ respectively, we have 
$\delta \leq \widetilde\delta = \lambda_{k+1}(\Lc) - \lambda_{k}(\Lc)$,
where $\widetilde\delta$ is defined in~\eqref{eq_eigengap_defn}.
Also, we can use Weyl's inequality to claim that
for all $i=1,\ldots,n$,
\begin{align*}
|\lambda_i(L) -\lambda_i(\Lc)| \leq \Vert L-\Lc\Vert_2 
\leq 12\sqrt{\frac{\ln n}{d}} < \frac{\delta}{2}
\end{align*}
if $\delta$ and $d$ satisfy the prescribed condition.
Thus
\begin{align*}
 \lambda_{k+1}(L) \geq \lambda_{k+1}(\Lc) -\frac{\delta}{2}
 = \lambda_{k}(\Lc) + \frac{\delta}{2} >0,
\end{align*}
which means $L$ has at most $k$ zero eigenvalues, \ie
at most $k$ connected components. 
Since, all nodes have positive degrees almost surely, hence,
every node corresponds to a connected component.
Due to third property of $L$, for every node, at least one of the $k$
leading eigenvectors has a non-zero component, and hence,
every row of $X$ is non-zero.
Thus, $\overline{X}$ is well-defined.

We now derive a perturbation bound involving $X$ and $\Xc$.
By Davis-Kahan $\sin\Theta$ theorem~\citep{Stewart_1990_book_AP},
we have if $2\Vert L-\Lc\Vert_2 < \delta$, then
\begin{align}
 \Vert \sin\Theta(X,\Xc)\Vert_2 \leq \frac{\Vert L - \Lc\Vert_2}{\delta}\;,
 \label{eq_sinTheta_bound}
\end{align}
where $\sin\Theta(X,\Xc)\in\R^{k\times k}$ is diagonal with entries same as
the sine of the canonical angles between the subspaces $X$ and $\Xc$. Let these angles be
denoted as
$\theta_1,\ldots,\theta_{k}\in[0,\frac{\pi}{2}]$ such that $\theta_1\geq \ldots\geq\theta_{k}$.
Then $\Vert \sin\Theta(X,\Xc)\Vert_2 = \sin \theta_1$.
On the other hand, one can see that the singular values 
for the matrix $X^T\Xc$ are given by $\cos\theta_1,\ldots\cos\theta_{k}$.
Thus, if $X^T\Xc = U_1\Sigma U_2^T$ is the singular value decomposition of $X^T\Xc$, then
\begin{align}
\Vert X - \Xc U_2U_1^T \Vert_F^2 
&= \textup{Trace}\left((X - \Xc U_2U_1^T)^T(X - \Xc U_2U_1^T)\right)
\nonumber
\\&= 2\textup{Trace}(I - U_1\Sigma U_1^T)
\nonumber
\\&= 2\sum_{i=1}^{k} (1-\cos\theta_i)
\leq 2\sum_{i=1}^{k} (1-\cos^2\theta_i)
\leq 2k\sin^2 \theta_1\;.
\label{eq_sinTheta_bound1}
\end{align}
From~\eqref{eq_sinTheta_bound} and~\eqref{eq_sinTheta_bound1},
we can conclude
that for $\delta>24\sqrt{\frac{\ln n}{d}}\geq2\Vert L-\Lc\Vert_2$,
\begin{align}
 \Vert X - Z(Z^TZ)^{-1/2}Q\Vert_F
 &= \Vert X - \Xc U_2 U_1^T\Vert_F
 \leq \sqrt{2k}\frac{\Vert L-\Lc\Vert_2}{\delta} \;,
\label{eq_sinTheta_bound2}
\end{align}
where $Q = UU_2U_1^T \in \R^{k\times k}$ is orthonormal.
One can see that $i^{th}$ row of $Z(Z^TZ)^{-1/2}Q$ is
$Z_{i\bull} (Z^TZ)^{-1/2}Q$, and the norm of $i^{th}$ row
is $(Z^TZ)_{\psi_i \psi_i}^{-1/2}$. Thus, on row normalization 
of this matrix, one obtains $ZQ$. Hence,
\begin{align*}
&\Vert \overline{X} - ZQ\Vert_F^2
= \sum_{i=1}^n \left\Vert \textstyle\frac{1}{\Vert X_{i\bull}\Vert_2} X_{i\bull}
- Z_{i\bull} Q\right\Vert_2^2
\\&= \sum_{i=1}^n \left\Vert 
\left(\textstyle\frac{1}{\Vert X_{i\bull}\Vert_2}  - (Z^TZ)_{\psi_i \psi_i}^{1/2}\right) X_{i\bull}
+ (Z^TZ)_{\psi_i \psi_i}^{1/2}\left(X_{i\bull} 
- Z_{i\bull} (Z^TZ)^{-1/2}Q\right)\right\Vert_2^2
\end{align*}
Now,
\begin{align*}
&\left\Vert 
\left(\textstyle\frac{1}{\Vert X_{i\bull}\Vert_2}  - (Z^TZ)_{\psi_i \psi_i}^{1/2}\right) X_{i\bull}
+ (Z^TZ)_{\psi_i \psi_i}^{1/2}\left(X_{i\bull} 
- Z_{i\bull} (Z^TZ)^{-1/2}Q\right)\right\Vert_2^2
\\&\leq \textstyle\sqrt{(Z^TZ)_{\psi_i \psi_i}}\left(
{\left| \Vert Z_{i\bull} (Z^TZ)^{-1/2}Q\Vert_2 - \Vert X_{i\bull} \Vert_2
\right|} + 
\Vert X_{i\bull} - Z_{i\bull} (Z^TZ)^{-1/2}Q\Vert_2\right)
\\&\leq 2\textstyle\sqrt{(Z^TZ)_{\psi_i \psi_i}}
\Vert X_{i\bull} - Z_{i\bull} (Z^TZ)^{-1/2}Q \Vert_2 \;.
\end{align*}
Substituting this bound above, we get
\begin{align*}
\Vert \overline{X} - ZQ\Vert_F^2
&\leq 4\sum_{i=1}^n (Z^TZ)_{\psi_i \psi_i}
\Vert X_{i\bull} - Z_{i\bull} (Z^TZ)^{-1/2}Q \Vert_2^2 
\\&\leq 4n_1 \Vert X - Z(Z^TZ)^{-1/2}Q \Vert_F^2\;,
\end{align*}
where $n_1 = \max_j (Z^TZ)_{jj}$ is the size of the largest partition
since we assumed that $n_1\geq \ldots \geq n_k$.
The bound in~\eqref{eq_perturbation_bound} follows by 
combining above bound with~\eqref{eq_sinTheta_bound2}
and Lemma~\ref{lem_random_deviation}.

\subsection{Proof of Lemma~\ref{lem_kmeans_condn}}
Let $\epsilon = (\ln n)^{-1/2}$.
From~\eqref{eq_perturbation_bound}, we have an upper bound on $\Vert \overline{X} - ZQ\Vert_F$
with probability $\left(1-O(n^{-2})\right)$. For convenience, let us denote this upper bound by 
$\beta$. The condition in~\eqref{eq_degree_condn} implies that
\begin{align*}
 \beta \leq 24\sqrt{\frac{2n_k}{C\ln n}} \leq \frac{\epsilon\sqrt{n_k}}{2}
\end{align*}
if $C$ is chosen sufficiently large. For large enough $n$, above inequality implies
\begin{align}
 \beta \leq \epsilon \left( \sqrt{n_k} - \beta\right),
 \label{eq_proof_kmeans_cond2}
\end{align}
which will be used to prove $\epsilon$-separability, \ie
$\eta_k(\overline{X}) \leq \epsilon \eta_{k-1}(\overline{X})$.
Since $ZQ$ has exactly $k$ distinct rows, we have
$\eta_k(\overline{X}) \leq \Vert X-ZQ\Vert_F \leq \beta$.
On the other hand, observe that all matrices in $\mathcal{M}_{n\times k}(r)$ have rank at most
$r$. Hence,
\begin{align*}
\eta_{k-1}(\overline{X}) = \min_{S\in\mathcal{M}_{n\times k}(k-1)} \Vert \overline{X} - S\Vert_F
\geq \min_{\text{rank}(S)\leq(k-1)} \Vert \overline{X} - S\Vert_F \;.
\end{align*}
It is well known that the minimum of the last quantity is $\lambda_k(\overline{X})$,
which is the smallest singular value of $\overline{X}$. Also 
Mirsky's theorem~\citep{Stewart_1990_book_AP} gives a bound on the perturbation of 
singular values, and hence, we have
\begin{align*}
 \left| \lambda_i(\overline{X}) - \lambda_i(ZQ) \right| \leq \Vert \overline{X} - ZQ \Vert_2 \leq\beta
\end{align*}
for $i=1,\ldots,k$.
Note here that the singular values of $ZQ$ are $\lambda_i(ZQ) = \sqrt{n_i}$, 
where the ordering of the values is due to our assumption $n_1\geq\ldots\geq n_k$.
From above arguments
\begin{align*}
 \eta_{k-1}(\overline{X}) \geq \lambda_k(\overline{X})
 \geq \left(\lambda_k(ZQ) - \beta\right) = \left(\sqrt{n_k} - \beta\right) \;.
\end{align*}
Hence, it follows that $\epsilon$-separability holds when~\eqref{eq_proof_kmeans_cond2} is satisfied, which is true under the assumption in~\eqref{eq_degree_condn}.

\subsection{Proof of Lemma~\ref{lem_kmeans}}
We observe from the proof of 
Lemma~\ref{lem_perturbation} that the $k$ distinct rows of $ZQ$,
form an orthonormal set of vectors in $\R^{k}$. Hence, for any $i,j\in\Vc$,
$\Vert Z_{i\bull}Q - Z_{j\bull}Q\Vert_2 = 0$ or $\sqrt{2}$,
where the former occurs if $Z_{i\bull}=Z_{j\bull}$, \ie $\psi_i=\psi_j$,
and the latter occurs if $\psi_i\neq\psi_j$. 
Now, consider $i,j\notin\Vc_{err}$. We have
$\Vert S^*_{i\bull} - Z_{i\bull}Q\Vert_2 < \frac{1}{\sqrt{2}}$,
$\Vert S^*_{j\bull} - Z_{j\bull}Q\Vert_2 < \frac{1}{\sqrt{2}}$, and hence,
\begin{align*}
\Vert Z_{i\bull}Q - Z_{j\bull}Q\Vert_2
&\leq \Vert S^*_{i\bull} - Z_{i\bull}Q\Vert_2 +
\Vert S^*_{j\bull} - Z_{j\bull}Q\Vert_2 + \Vert S^*_{i\bull} - S^*_{i\bull}\Vert_2
<
\sqrt{2}
\end{align*}
whenever $S^*_{i\bull} = S^*_{j\bull}$. So, from the previous observation, 
$\Vert Z_{i\bull}Q - Z_{j\bull}Q\Vert_2 = 0$, \ie $\psi_i = \psi_j$,
which proves the first claim.

The second claim follows by using arguments similar to~\citep{Rohe_2011_jour_AnnStat}.
Note that for all $i\in\Vc_{err}$, we have $2\Vert S^*_{i\bull} - Z_{i\bull}Q\Vert_2^2 \geq1$.
Therefore,
\begin{align}
 |\Vc_{err}| = \sum_{i\in\Vc_{err}} 1
 \leq 2\sum_{i\in\Vc_{err}} \Vert S^*_{i\bull} - Z_{i\bull}Q\Vert_2^2
 \leq 2\Vert S^* - ZQ\Vert_F^2 \;.
 \label{eq_W_bound}
\end{align}
Since, $S^*$ is a sub-optimal solution satisfying~\eqref{eq_suboptimal_kmeans},
we can write
$\Vert \overline{X} - S^*\Vert_F \leq \gamma \Vert \overline{X} - S\Vert_F$,
for all $S\in\mathcal{M}_{n\times k}(k)$. In particular, $ZQ\in\mathcal{M}_{n\times k}(k)$ and so
$\Vert \overline{X} - S^*\Vert_F \leq \gamma \Vert \overline{X} - ZQ\Vert_F$.
Hence,
\begin{align*}
 \Vert S^* - ZQ\Vert_F \leq \Vert \overline{X} - ZQ\Vert_F + \Vert \overline{X} - S^*\Vert_F
 \leq (1+\gamma) \Vert \overline{X} - ZQ\Vert_F\;.
\end{align*}
The lemma follows by combining above inequality with~\eqref{eq_W_bound},
and using the relation $(1+\gamma)^2 \leq 2(1+\gamma^2)$.

\section{Proofs of Corollaries in Section~\ref{sec_specialcases}}

\subsection{Proof of Corollary~\ref{cor_equal_size_uniform}}
We observe that for the specified model,
the matrix $G\in\R^{k\times k}$, as defined in Lemma~\ref{lem_expected1}, is given by
\begin{align*}
 G_{ij} = \left\{\begin{array}{ll}
 \displaystyle\frac{p\alpha_{r,n}}{r}\binom{\frac{n}{k}-2}{r-2} + \frac{q\alpha_{r,n}}{r}\binom{n-2}{r-2}
 & \text{if } i = j,
 \\\\
 \displaystyle\frac{q\alpha_{r,n}}{r}\binom{n-2}{r-2} & \text{if } i\neq j.
                 \end{array}\right.
\end{align*}
Thus, $G$ is of the form $G = aI + b\mathbf{1}$, where $\mathbf{1}$ is constant matrix of ones.
It is easy to verify that for such a matrix, the minimum eigenvalue is $a$.
Hence, we have $\lambda_{\min}(G) = \frac{p\alpha_{r,n}}{r}\binom{\frac{n}{k}-2}{r-2}$.
Also, we can compute 
\begin{align}
 d = \left(p\alpha_{r,n}\binom{\frac{n}{k}-1}{r-1} + q\alpha_{r,n}\binom{n-1}{r-1}\right)
 > q\alpha_{r,n}\binom{n-1}{r-1}\;.
 \label{eq_Dmin_bound_uniform}
\end{align}
Note that since the partitions are balanced and node degrees behave identically,
the second term in~\eqref{eq_delta_charac} is zero, and
we can compute ${\delta}$ as
\begin{align}
\delta=\frac{{n}p\alpha_{r,n}}{{kd}r}\binom{\frac{n}{k}-2}{r-2} 
\geq \frac{n}{k r} \frac{p\alpha_{r,n}\binom{\frac{n}{k}-2}{r-2}}{(p+q)\alpha_{r,n}\binom{n-1}{r-1}}
\geq\frac{p(r-1)}{r(p+q)}\frac{k\binom{n/k}{r}}{\binom{n}{r}}\;,
 \label{eq_delta_bound_uniform}
\end{align}
where the first inequality follows by observing that 
$d \leq (p+q)\alpha_{r,n}\binom{n-1}{r-1}$. Now, observe
that under the given condition on $\alpha_{r,n}$, we have
\begin{align*}
 \delta^2d 
 \geq C'\frac{\alpha_{r,n}}{n}\binom{n}{r}\left(\frac{k\binom{n/k}{r}}{\binom{n}{r}}\right)^2
 \geq C''k^{2r-1}(\ln n)^2 \left(\frac{k(\frac{n}{k})^r}{n^r}\right)^2 
 \geq C'' k(\ln n)^2,
\end{align*}
where $C''$ is a constant depending only on $C,p,q$ and $r$, that is obtained 
from \eqref{eq_Dmin_bound_uniform} and \eqref{eq_delta_bound_uniform}.
The last inequality uses the relation $\frac{a^b}{4(b!)}\leq \binom{a}{b}\leq \frac{a^b}{b!}$.
Choosing $C$ sufficiently large, the condition of Theorem~\ref{thm_main_consistency}
is satisfied, and hence, we can conclude from
Theorem~\ref{thm_main_consistency} that $\err =
O\left(\frac{n\ln n}{\delta^2d}\right)$, which simplifies to the
stated claim.

\subsection{Proof of Corollary~\ref{cor_equal_size_nonuniform}} 
The proof is quite similar to that of Corollary~\ref{cor_equal_size_uniform}
since the matrices $G$ and $\Dc$ are linear combinations of the corresponding matrices
for the $m$-uniform hypergraphs. We compute
\begin{align*}
\lambda_{\min}(G) = \sum\limits_{m=2}^{M} \frac{p\alpha_{m,n}}{m}\binom{\frac{n}{k}-2}{m-2} 
\geq \frac{p\alpha_{r,n}}{r}\binom{\frac{n}{k}-2}{r-2}  \;,
\end{align*}
where $\theta_r>0$ and we ignore all terms for $m>r$. 
Substituting the relation for $\alpha_{r,n}$, we can write
\begin{align*}
\lambda_{\min}(G) \geq C_1\frac{\theta_r n^{a-2} (\ln n)^b}{k^{r-2}} 
\end{align*} 
for some constant $C_1$ depending on $p,q$ and $r$. Also,
\begin{align*}
 d &= \sum_{m=2}^{M}\left(p\alpha_{m,n}\binom{\frac{n}{k}-1}{m-1} + q\alpha_{m,n}\binom{n-1}{m-1}\right)
 \\&> q\sum_{m=2}^{M}\frac{m}{n}\alpha_{m,n}\binom{n}{m}
 \geq n^{a-1} (\ln n)^b  q\sum_{m=2}^{M} m\theta_m \;.
\end{align*}
Similarly, one can verify that
$d \leq (p+q)n^{a-1}(\ln n)^b\sum\limits_{m=2}^{M}m\theta_m$. Thus,
\begin{align}
 \delta^2d &\geq \frac{n^2(\lambda_{\min}(G))^2}{k^2d}
 \geq \frac{C_1^2\theta_r^2}{(p+q)} \frac{n^{a-1}(\ln n)^b}{k^{2r-2} \sum\limits_{m=2}^{M} m\theta_m} \;,
 \label{eq_delta_bound_nonuniform}
\end{align}
where the inequality follows from above bounds on $\lambda_{\min}(G)$ and $d$.
Under the condition~\eqref{eq_nonuniform_balanced_condn} with large enough $C$, 
the condition of Theorem~\ref{thm_main_consistency} holds and the claim follows.

\subsection{Proofs for Lemmas~\ref{lem_3uniform} and~\ref{lem_23nonuniform}}
As in the previous corollaries, we can say that, for both cases, the second term in~\eqref{eq_delta_charac}
is zero. Hence, $\delta>0$ if and only if $\lambda_{\min}(G)>0$.
For the setting of Lemma~\ref{lem_3uniform}, one can compute
\begin{align*}
 G_{ij} = \left\{\begin{array}{ll}
 \displaystyle\frac{\alpha_{3,n}}{3}
 \left[p_1\left(\frac{n}{k}-2\right) + p_2\left(n-\frac{n}{k}\right)\right]
 & \text{if } i = j,
 \\\\
 \displaystyle\frac{\alpha_{3,n}}{3}
 \left[p_2\left(\frac{2n}{k}-2\right) + p_3\left(n-\frac{2n}{k}\right)\right]
 & \text{if } i\neq j.
                 \end{array}\right.
\end{align*}
Using an observation made in the proof of Corollary~\ref{cor_equal_size_uniform},
we have $\lambda_{\min}(G) = G_{11} - G_{12}$,
and~\eqref{eq_3uniform_condition} is equivalent to stating $G_{11} > G_{12}$.

The same arguments are valid for Lemma~\ref{lem_23nonuniform}, where $G$ is of the form
\begin{align*}
 G_{ij} = \left\{\begin{array}{ll}
 \displaystyle\frac12 + \displaystyle\frac{\alpha_{3,n}}{3}
 \left[p_1\left(\frac{n}{k}-2\right) + p_2\left(n-\frac{n}{k}\right)\right]
 & \text{if } i = j,
 \\\\
 \displaystyle\frac{\alpha_{3,n}}{3}
 \left[p_2\left(\frac{2n}{k}-2\right) + p_3\left(n-\frac{2n}{k}\right)\right]
 & \text{if } i\neq j.
                 \end{array}\right.
\end{align*}

\subsection{Proof of Lemma~\ref{lem_plantedclique}}

As mentioned in Lemma~\ref{lem_expected1}, one can write $\Ac$ as
$\Ac = ZGZ^T - J$. In the present case, $G\in\R^{2\times2}$ is given by
\begin{align*}
 G_{ij} = \left\{\begin{array}{ll}
 \displaystyle\frac{1}{2r}\binom{s-2}{r-2} + \frac{1}{2r}\binom{n-2}{r-2}
 & \text{if } i = j=1,
 \\\\
 \displaystyle\frac{1}{2r}\binom{n-2}{r-2}
 & \text{otherwise}.
                 \end{array}\right.
\end{align*}
One can also verify that the diagonal matrices $\Dc$ and $J$ are given by
\begin{align*}
 \Dc_{ii} = \left\{\begin{array}{ll}
 \displaystyle\frac{1}{2}\binom{s-1}{r-1} + \frac{1}{2}\binom{n-1}{r-1}
 & \text{if } i \in s \text{-clique},
 \\\\
 \displaystyle\frac{1}{2}\binom{n-1}{r-1}
 & \text{otherwise}.
                 \end{array}\right.
\end{align*}
and
\begin{align*}
 J_{ii} = \left\{\begin{array}{ll}
 -\displaystyle\frac{1}{2r}\binom{s-2}{r-1} - \frac{1}{2r}\binom{n-2}{r-1}
 & \text{if } i \in s \text{-clique},
 \\\\
 -\displaystyle\frac{1}{2r}\binom{n-2}{r-1}
 & \text{otherwise}.
                 \end{array}\right.
\end{align*}
Hence, the claim follows by substituting above relations in~\eqref{eq_delta_charac}.
It is more convenient to write~\eqref{eq_delta_charac} using the notations $\Dt$ and $\Jt$ defined
in the proof of Lemma~\ref{lem_expected}, and it can be written as
\begin{align}
 \delta = \frac{s\lambda_{\min}(G)}{\Dt_{11}} - \left| \frac{\Jt_{11}}{\Dt_{11}} - \frac{\Jt_{22}}{\Dt_{22}} \right|,
 \label{eq_delta_plantedclique}
\end{align}
where we use the fact $s< (n-s)$. Note that $G$ has non-negative eigenvalues,
and we can use the following  relation for $2\times2$ matrices
\begin{align*}
\lambda_{\min}(G) \geq \frac{\text{det}(G)}{\text{Trace}(G)}
= \frac{\frac{1}{2r}\binom{s-2}{r-2}\binom{n-2}{r-2}}{\binom{s-2}{r-2}+2\binom{n-2}{r-2}}
\geq \frac{1}{6r}\binom{s-2}{r-2}.
\end{align*}
Combining this with the observation $\Dt_{11} \leq \binom{n-1}{r-1}$, we can argue that
the first term in~\eqref{eq_delta_plantedclique} is at least
\begin{align*}
 \frac{s\binom{s-2}{r-2}}{6r\binom{n-1}{r-1}} \geq C_1 \left(\frac{s}{n}\right)^{r-1}
\end{align*}
for some $C_1>0$.
On the hand, the second term in~\eqref{eq_delta_plantedclique} can be computed as
\begin{align*}
\left| \frac{\Jt_{11}}{\Dt_{11}} - \frac{\Jt_{22}}{\Dt_{22}} \right| 
&= \left| \frac{ \binom{s-1}{r-1}\binom{n-2}{r-1} - \binom{n-1}{r-1}\binom{s-2}{r-1}}{
\binom{n-1}{r-1} \left[\binom{n-1}{r-1} + \binom{s-1}{r-1}\right]}\right|
\\& \leq \frac{\binom{s-1}{r-1}}{\binom{n-1}{r-1}} \left( \frac{n-r}{n-1} - \frac{s-r}{s-1}\right)
\\&\leq \frac{\binom{s-1}{r-1}}{\binom{n-1}{r-1}} \left(\frac{r-1}{s-1}\right) 
\leq \frac{C_2}{s}\left(\frac{s}{n}\right)^{r-1}
\end{align*}  
for some $C_2>0$. From above, one can conclude that $\delta>0$ when 
$C_1 > \frac{C_2}{s}$, where both $C_1$ and $C_2$ depend only on $r$. Hence, the claim.

\footnotesize

\end{document}